\documentclass[10pt,twocolumn,letterpaper]{article}

\usepackage{ICCV}              % To produce the CAMERA-READY version
\usepackage[accsupp]{axessibility}
\usepackage{graphicx}
\usepackage{amsmath}
\usepackage{amssymb}
\usepackage{booktabs}
\usepackage{csquotes}
\usepackage{newtxtext,newtxmath}
\usepackage{enumitem}
\setlist[itemize]{noitemsep, topsep=0pt, left=0pt}
\usepackage{mathrsfs}
\usepackage{colortbl}
\usepackage[cal=cm]{mathalfa}
\usepackage{algorithm}
\usepackage{algorithmic}

\usepackage{multirow}
\usepackage{epsfig}
\usepackage{relsize}

\definecolor{iccvblue}{rgb}{0.21,0.49,0.74}
\usepackage[pagebackref,breaklinks,colorlinks,allcolors=iccvblue]{hyperref}

\usepackage[capitalize]{cleveref}
\crefname{section}{Sec.}{Secs.}
\Crefname{section}{Section}{Sections}
\Crefname{table}{Table}{Tables}
\crefname{table}{Tab.}{Tabs.}

\def\paperID{6081} % *** Enter the ICCV Paper ID here
\def\confName{ICCV}
\def\confYear{2025}

\begin{document}

%%%%%%%%% TITLE - PLEASE UPDATE
\title{UPRE: Zero-Shot Domain Adaptation for Object Detection via Unified Prompt and Representation Enhancement
}

\author{Xiao Zhang\textsuperscript{1,2}\thanks{Work partially done during the internship at AMAP, Alibaba Group.},~Fei Wei\textsuperscript{2},~Yong Wang\textsuperscript{2},~Wenda Zhao\textsuperscript{1}\thanks{Corresponding author.},
~Feiyi Li\textsuperscript{1},~Xiangxiang Chu\textsuperscript{2} \\
\textsuperscript{1}Dalian University of Technology \quad \textsuperscript{2}AMAP, Alibaba Group \\
{\tt\small \{1204855526@mail.dlut.edu.cn, zhaowenda@dlut.edu.cn, lifeiyi@mail.dlut.edu.cn\}} \\
{\tt\small \{xixia.wf, wangyong.lz, chuxiangxiang.cxx\}@alibaba-inc.com}
}
\maketitle

%%%%%%%%% ABSTRACT
\begin{abstract}
Zero-shot domain adaptation (ZSDA) presents substantial challenges due to the lack of images in the target domain. Previous approaches leverage Vision-Language Models (VLMs) to tackle this challenge, exploiting their zero-shot learning capabilities. However, these methods primarily address domain distribution shifts and overlook the misalignment between the detection task and VLMs, which rely on manually crafted prompts. To overcome these limitations, we propose the unified prompt and representation enhancement (UPRE) framework, which jointly optimizes both textual prompts and visual representations. Specifically, our approach introduces a multi-view domain prompt that combines linguistic domain priors with detection-specific knowledge, and a visual representation enhancement module that produces domain style variations. Furthermore, we introduce multi-level enhancement strategies, including relative domain distance and positive-negative separation, which align multi-modal representations at the image level and capture diverse visual representations at the instance level, respectively. Extensive experiments conducted on nine benchmark datasets demonstrate the superior performance of our framework in ZSDA detection scenarios.  Code is available at https://github.com/AMAP-ML/UPRE.
\end{abstract}

\section{Introduction}
\label{sec:intro}
\begin{figure}[t]
  \setlength{\abovecaptionskip}{0.2cm}
  \setlength{\belowcaptionskip}{-0.4cm}
  \centering
   \includegraphics[width=0.95\linewidth]{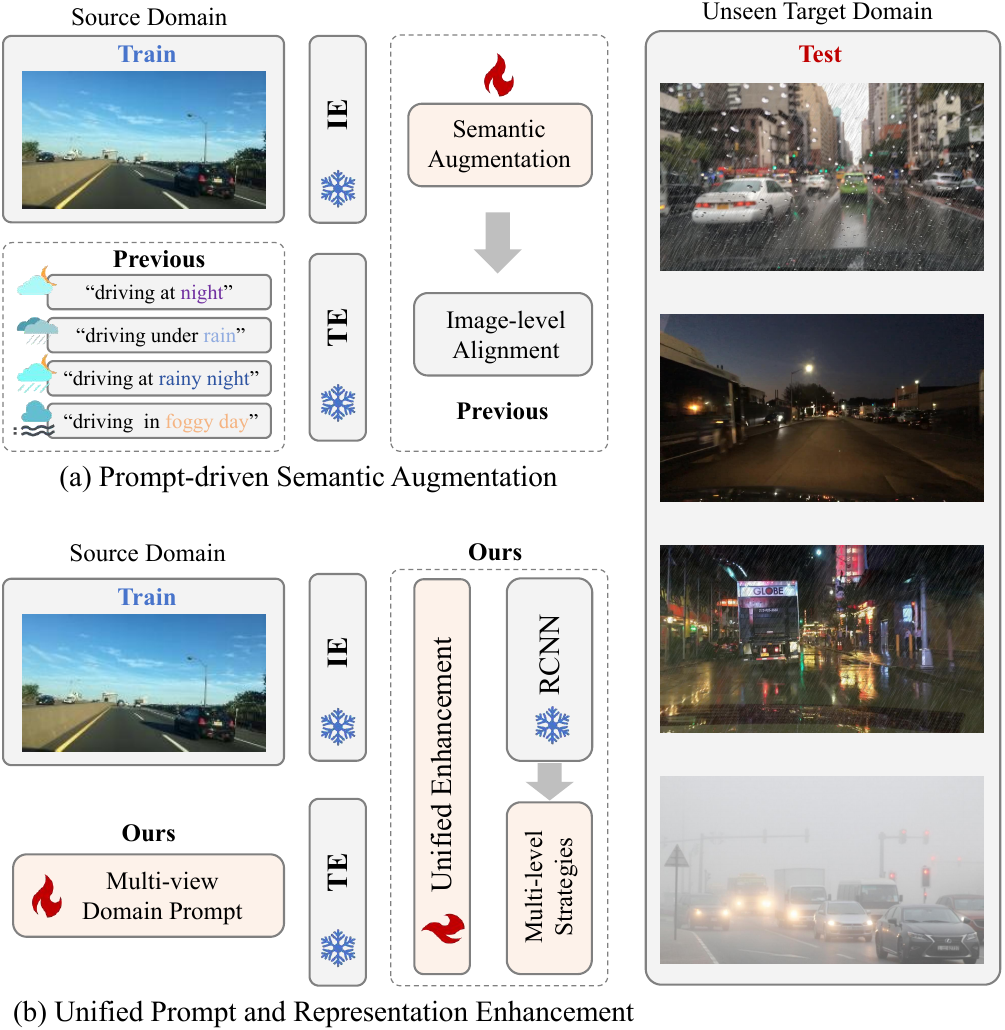}
   \caption{
   \textbf{(a)} Prompt-driven semantic augmentation generates pseudo target domain features with manually crafted prompts.
   \textbf{(b)} The proposed unified prompt
and representation enhancement creates the synthetic target domain representations via learnable prompts. Learnable prompt and unified enhancement jointly mitigate detection and domain biases. IE represents the image encoder, and TE denotes the text encoder.}
   \label{fig:2}
\end{figure}

Domain adaptation has gained widespread attention in recent years \cite{hsu2020progressive,belal2024multi,mattolin2023confmix,vs2023instance,oza2023unsupervised,ngo2024learning,li2024comprehensive,li2024ada}. However, obtaining image priors, even unlabeled, is not always feasible, thereby limiting their applicability in real-world scenarios. Zero-shot domain adaptation (ZSDA) \cite{min2020domain,xia2020hgnet,wang2021domain,fahes2023poda,yang2024unified,azuma2024zodi} aims to address this constraint by enabling adaptation to target domains without prior image exposure. The development of vision-language models (VLMs) \cite{li2021align,li2022blip,alayrac2022flamingo,kim2021vilt,chu2025gpg,wang2022ofa,chen2022pali,li2025next} has markedly advanced ZSDA, by utilizing textual prompts to describe unseen domains  through their inherent zero-shot capabilities \cite{fahes2023poda,yang2024unified}. Despite these advancements, our analysis reveals two primary limitations in existing VLM-based object detection approaches: 1) \textit{domain bias}, where distribution shifts between source and target domains introduces task-agnostic noise, impairing model performance; 2) \textit{detection bias}, where models such as CLIP \cite{radford2021learning} emphasize global image representations but overlook instance-level details crucial for precise object localization. This deficiency stems from manually constructed prompts, which inadequately capture the contextual attributes of foreground and background objects.

Previous studies \cite{vidit2023clip,fahes2023poda,yang2024unified} have explored the mitigation of domain bias through prompt-driven semantic augmentation, as illustrated in Fig.~\ref{fig:2}(a). These methods utilize manually crafted prompts combined with image-level alignment to produce the representations of pseudo target domain. Despite their effectiveness in addressing domain bias, these approaches overlook and intensify detection bias during the fine-tuning process. Conversely, methods  \cite{du2022learning,he2023unsupervised,zhao2024scene,li2024learning} aimed at reducing detection bias tend to reinforce domain bias. They endeavors to minimize detection bias by learning prompt representations. However, in the absence of target domain images, the training of prompt representation is restricted to the source domain. Consequently, while these prompts effectively encapsulate representations for the detection task within the source domain, they unintentionally magnify domain bias. This gap highlights the pressing need for method that can simultaneously address both domain and detection biases.

In this paper, we propose the unified prompt and representation enhancement (UPRE) framework, a novel approach that jointly enhances textual prompt and visual representations. As illustrated in Fig.~\ref{fig:2}(b), our method establishes a cooperative relationship between prompt optimization and visual representation learning. Specifically, we design a multi-view domain prompt (MDP) that provides language-modal priors for the target domain while capturing diverse adaptation knowledge essential for cross-domain object detection. Furthermore, we introduce unified representation enhancement (URE), a shared module between both vision and language modalities designed to generate target domain representations from source domain data. The URE enhances the diversity of the domain styles under the guidance of prompt representations, effectively alleviating domain bias. Concurrently, the learnable prompt representations of MDP utilize the target domain representations generated by URE to acquire instance-level knowledge for object detection, thereby reducing detection bias.

To further enhance this interaction, we introduce two novel enhancement strategies: relative domain distance (RDD) and positive-negative separation (PNS), which together form a multi-level training framework. RDD aligns multi-modal representations at the image level to facilitate adaptation, while PNS leverages background context from negative proposals and object information from positive ones to capture diverse instance knowledge. Through the unified training process, our approach enhances both prompt and visual representations, thereby effectively mitigating detection and domain biases. This leads to notable improvements in adaptation and detection capabilities, empowering object detectors to accurately adapt to previously unseen domains. Our proposed method demonstrates superior performance across three domain adaptation tasks on nine datasets, including adverse weather conditions, cross-city scenarios, and virtual-to-real transitions.

In summary, our contributions are as follows:
\begin{itemize}
\item We identify and tackle the challenges of detection and domain biases in VLMs for ZSDA by introducing a unified prompt and representation enhancement (UPRE) approach tailored for object detection.
\item We introduce the MDP mechanism to learn prompt representations tailored for cross-domain detection, and propose the URE to increase the diversity of domain styles.
 
\item We develop the RDD technique to enable effective image-level adaptation, and the PNS to enhance the capability of precise localization at the instance level.
\end{itemize}

\section{Related Work}
\subsection{Zero-shot Domain Adaptation}
ZSDA~\cite{peng2018zero, du2024boosting,wang2019conditional,wang2020adversarial,luo2023similarity,wang2021learning,lengyel2021zero,kutbi2021zero,fan2024rapid} presents a challenging learning paradigm that transfers knowledge from a source domain to a target domain without accessing any target domain data. Min \textit{et al.}~\cite{min2020domain} address generalized zero-shot learning by eliminating domain-specific visual biases. Wang \textit{et al.}~\cite{wang2021domain} introduce a domain shift preservation method, which relies on GAN~\cite{goodfellow2014generative} to maintain domain invariance during adaptation. Recent advances have explored VLMs to improve adaptation performance. Fahes \textit{et al.}~\cite{fahes2023poda} propose a prompt-driven approach to guide domain adaptation. Yang \textit{et al.}~\cite{yang2024unified} also propose a unified language-driven method to bridge the gap between the source and target domains. In parallel, Single Domain Generalization~\cite{wu2022single,rao2023srcd,danish2024improving,vidit2023clip,liu2024unbiased,lee2024object,wu2024g}, focuses on generalizing models to multiple unseen target domains simultaneously. Lee \textit{et al.}~\cite{lee2024object} propose an object-aware domain generalization method to address misclassification and imprecise localization challenges. Subsequently, Liu \textit{et al.}~\cite{liu2024unbiased} introduce a structural causal model to analyze data and feature biases in the task. While recent works~\cite{vidit2023clip,fahes2023poda,yang2024unified} leverage the zero-shot learning capabilities of VLMs to mitigate domain bias, they overlook downstream task biases, particularly detection bias, which remain underexplored for ZSDA.

\begin{figure*}
  \centering
  \setlength{\abovecaptionskip}{0.2cm}
  \setlength{\belowcaptionskip}{-0.2cm}
   \includegraphics[width=0.90\linewidth]{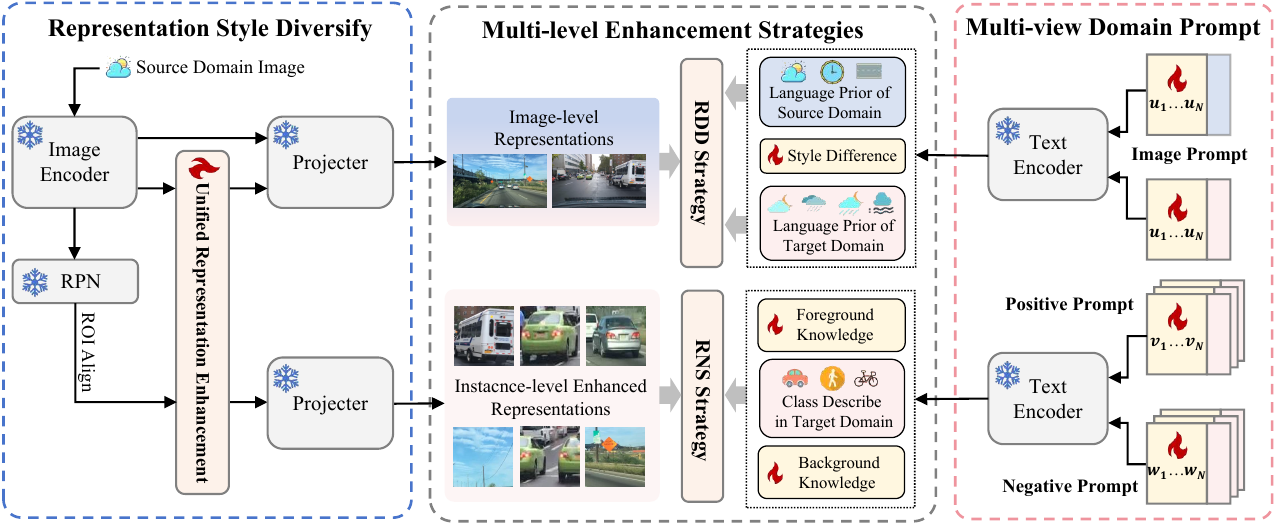}
   \caption{\textbf{Overview of unified prompt
and representation enhancement}. Initially, the multi-view domain prompt provides language prior while also learns adaptation knowledge for object detection.  
Meanwhile, unified representation enhancement enriches source domain features with RPN, generating pseudo target domain representations.
Finally, two enhancement strategies constrain language and visual representations, training the unified framework effectively.
The symbols 
\makebox[5pt][l]{\includegraphics[height=0.8em]{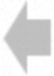}} represent losses.
   }
   \label{fig:3}
\end{figure*}

\subsection{Domain Prompt Learning}
Recent advances in domain prompt learning \cite{ge2023domain,xin2024mmap,li2024learning,bose2024stylip,zhao2024learning,singha2023ad,bai2024diprompt,wang2024pond,cheng2024disentangled} have demonstrated promising capabilities for cross-domain knowledge transfer through VLMs with minimal intervention. Existing approaches can be broadly categorized into two paradigms: domain-specific adaptation and doamin-invariant generalization. Domain-specific methods customizes prompts to capture domain related characteristic. For instance, Ge \textit{et al.} \cite{ge2023domain} adapt domain-aware prompts to bridge domain gaps, while Cao \textit{et al.} \cite{cao2024domain} dynamically adjust prompts based on domain features. Li \textit{et al.} \cite{li2024learning} enhance detection performance by incorporating prompt tuning in the detection head. Conversely, domain-invariant approaches focus on learning generalized prompts for unseen domains. Zhao \textit{et al.} \cite{zhao2024learning} focus on learning domain-invariant prompts for VLMs, utilizing domain-agnostic representations to enhance adaptability. Singha \textit{et al.} \cite{singha2023ad} introduce AD-CLIP, adapting prompts within CLIP \cite{radford2021learning} representations to promote generalization. However, these methods face inherent limitations in ZSDA scenarios. Domain-specific approaches struggle to mitigate domain bias without image priors, while domain-invariant methods often decrease task-specific discriminability. The core challenge lies in effectively leveraging limited source-domain information to address both domain shift and downstream task misalignment. In this paper, we propose a novel approach that unified prompt learning and representation enhancement to address this challenge.

\section{Method}
To jointly address detection and domain biases in zero-shot domain adaptation, we propose a novel framework that cooperatively integrates prompt optimization with cross-domain representation synthesis. As depicted in Fig.~\ref{fig:3}, our approach establishes a unified learning system where domain-aware prompt representation and visual representation enhancement mutually reinforce each other.
Our framework comprises three main components: 
\begin{itemize}
\item Multi-view Domain Prompt (MDP): a hybrid prompts combining linguistic knowledge of target domain priors (image/instance-level) with learnable prompts to learn cross-domain knowledge for object detection.
\item Unified Representation Enhancement (URE): a representation style diversify module through feature transformation and multi-modal projection.  
\item Multi-level Enhancement Strategies: incorporating Relative Domain Distance (RDD) and Positive-Negative Separation (PNS) for representation alignment between linguistic and visual modals. 
\end{itemize}

\paragraph{Problem Formulation. \label{sec.3.1}}
Traditional object detection dataset is defined as $\mathcal{D} = \{(\mathcal{X}_n,\mathcal{O}_n)\}_{n=1}^{|\mathcal{D}|}$, where $\mathcal{X}_n$ is the $n$-th image,
$\mathcal{O}_n = \{o_{nm}\}_{m=1}^{|\mathcal{O}_n|}$ is the corresponding set of annotated objects, and $m$ denotes the $m$-th objects. Each object $o$ is a pair of object bounding box $b \in \mathbb{R}^4$ and category label $y_c \in \mathcal{C}$, where $\mathcal{C}$ is the category space of the dataset. Zero-shot domain adaptation for object detection aims to detect objects in an unseen target domain using a model trained exclusively on the source domain. Specifically, the dataset space $\mathcal{D}$ is divided into source
domain data $\mathcal{D}^s =\{(\mathcal{X}_n,\mathcal{O}_n)\}_{n=1}^{|\mathcal{D}^s|} $ and unseen target domain data $\mathcal{D}^t \!=\! \{\mathcal{X}_n\}_{n=1}^{|\mathcal{D}^t|}$,
where $\mathcal{D} \!=\! \mathcal{D}^s \cup \mathcal{D}^t$ and $\mathcal{D}^s \cap \mathcal{D}^t = \varnothing$. For each category $c \in \mathcal{C} $, we use a pretrained text encoder $\mathcal{T}$ to encode its semantic embedding as: $\textbf{t}_c = \mathcal{T}(\mathcal{R}(c))$, where $\mathcal{R}(c)$ is the textual prompt of class $c$, e.g. a photo of a $[class]$. Besides, we define a class of \enquote{\scalebox{0.8}{$\mathcal{C}_{bg}$}} to represent the background category. Given an image, the
image encoder $\mathcal{I}$ of CLIP extracts a set of hierarchical feature maps, then the RPN generates a set of proposals $\mathcal{P}$. Subsequently, we perform ROI-Align to extract proposal features $\{F_p\}~ p\in \mathcal{P}$, and feed it into the projector (the rest layer of CLIP) to extract proposal representations as: $\textbf{e}_p = \mathcal{I}(F_p)$. The probability of a proposal belonging to category $c$ is computed as:
\begin{equation}
P_c = \frac{\exp(f( \textbf{e}_p, \textbf{t}_c ))}{\sum_{c' \in \mathcal{C} \cup \{ \textsmaller{\mathcal{C}_{bg}}\}} \exp(f( \textbf{e}_p, \textbf{t}_{c'} ))}
     \label{eq.3}
\end{equation}
where $f( \cdot,\cdot )$ denotes the cosine similarity function.

\subsection{Multi-view Domain Prompt \label{sec.3.2}}  
Previous methods \cite{cao2024domain, li2024learning, zhao2024learning} rely on target domain images and assume that learnable prompts can fully replace manually designed prompts.
In our approach, we retain the static human-defined prompt and prepend it with learnable prompts.
By retaining the original human-defined context, our design allows the learnable prompt to concentrate on capturing multi-view knowledge. 
Specifically, we propose MDP to learn both domain adaptation knowledge for bridging domain gaps and object localization information for accurate dense predictions.
As illustrated in Fig.~\ref{fig:3} (right), MDP consists of three parts, including image, positive, and negative prompts. 
The image prompt provides language priors of target domain such as lighting and visibility, while further learning the style differences between domains at the image level.
Initially, we define $\{u_l,v_l,w_l\}^L_{l=1}$ as the learnable context vectors with same dimension.
The image prompt representation $\mathcal{R}_i^d$ for given
domain $d \subset \{s,t\}$ is
defined as:
\begin{equation}
\mathcal{R}_i^d = \big[ u_1, u_2, \dots, u_L, k_d\big],
\end{equation}
where $k_d$ is the word representations of
\enquote{a photo taken on a $[domain]$}.
Positive proposals encode the characteristics of foreground objects, whereas negative proposals focus on capturing the contextual information in background.
To model the stylistic variations of foreground objects in the target domain, we introduce the positive prompt representation $\mathcal{R}_p$.
Given target domain $t$ and class $c$, positive prompt representation $\mathcal{R}_p^t(c)$ is defined as:
\begin{equation}
\mathcal{R}_p^t(c) = \big[v_1, v_2 \dots v_L, (k_t , k_c)\big],
\end{equation}
where $(k_t,k_c)$ denotes the word representations of \enquote{a $[domain]$ photo of a $[class]$}.
Similarly, the negative prompt representation $\mathcal{R}_n^t(\mathcal{C}_{\text{bg}})$, which is designed to capture contextual background knowledge, is defined as:
\begin{equation}
\mathcal{R}_n^t(\textsmaller{\mathcal{C}_{bg}}) = \big[w_1, w_2 \dots w_L, (k_t, k_{\textsmaller{\mathcal{C}_{bg}}})\big],
\end{equation}
where $( k_t,k_{\textsmaller{\mathcal{C}_{bg}}} )$ is the word representations of \enquote{a $[domain]$ photo of an $[unknown \;\, class]$}.

\subsection{Unified Representation Enhancement}
Despite lacking prior knowledge of target domain images, VLMs \cite{wangsimvlm,yu2022coca,radford2021learning,li2022grounded} can still describe unseen domains through prompts.
CLIP-GAP \cite{vidit2023clip} exploit it to train semantic augmentation for domain transformation but overlooks the deviation term.
Later, PODA \cite{fahes2023poda} addresses this by introducing AdaIN \cite{huang2017arbitrary} to globally transfer feature styles.
However, the style of an image often varies dynamically in different regions of the real world.
For example, for a rainy and night image, nearby regions may reflect a style of ``rainy''  due to brighter lighting, while distant regions may reflect a ``night'' style due to insufficient lighting.
Thus, basic style transform approach using the global style parameters is inadequate for complex domain adaptation detection task.

Subsequently, we propose the learnable mean enhancement $\mathcal{E}_\mu \in \mathbb{R}^{C \times M \times N}$ and learnable deviation enhancement $\mathcal{E}_\sigma \in \mathbb{R}^{C \times M \times N}$ to establish a fine-grained solution.
Specifically, we align the pseudo target domain features $F_{s \to t}$, which are generated by enhancing the source image features $F_s$ through URE, with the linguistic counterparts of the target image features $F_t$.
To generate pseudo target domain features $F_{s \to t}$, we first segment $F_s$ into $M \times N$ patches.
After that, we apply $\mathcal{E}_\mu$ and $\mathcal{E}_\sigma$ to enhance the style-specific elements of $F_s$ as follows:
\begin{equation}
F_{s \to t} = {\left\{ \mathcal{E}_\sigma^{j} \cdot F_s^{j} + \mathcal{E}_\mu^{j}\right\} }_{j=1}^{M \times N},
\end{equation}
where $j$ represents the $j$-th patch.
The source image features $F_s$ are enhanced to pseudo target features $F_{s \to t}$, capturing the stylistic characteristics of the target domain in real-world scenarios.

\subsection{Multi-level Enhancement Strategies\label{sec.3.3}}
To further unlock the potential of VLMs, we propose RDD and PNS, which enable prompt and enhancement training at both the image and instance levels.

\textbf{Relative Domain Distance.}
We introduce an image-level enhancement strategy to better describe the distribution of the target domain.
Specifically, we feed $\mathcal{R}_i^d$ into text encoder $\mathcal{T}$ to generate text embedding $\textbf{t}_i^s$ and $\textbf{t}_i^t$.
The image representations $\textbf{e}_i^{s \to t}$ and $\textbf{e}_i^s$ are computed by feeding $F_{s \to t}$ and $F_s$ to an image encoder $\mathcal{I}$.
Then, the distance between image prompt representations and the enhanced representations is pushed closer by a loss $\mathcal{L}_{a}$:
\begin{equation}
\mathcal{L}_{a} = \mathbb{E}\left[1 - f (\textbf{e}_i^{s \to t} , \textbf{t}_i^t)\right],
\end{equation}
It is crucial to regulate URE to ensure that it captures the style of target domain  while maintaining the integrity of semantic information for objects. 
Thus, the constraint loss $\mathcal{L}_s$ is defined as:
\begin{equation}
\mathcal{L}_s = \mathbb{E}\left[\|\textbf{e}_i^s- \textbf{e}_i^{s \to t}\|_1\right],
\end{equation}
where $\| \cdot \|_1$ represents $\mathcal{L}1$ distance metric.
Besides, to efficiently learn pseudo target domain representations, we introduce a relative domain distance loss $\mathcal{L}_r$:
\begin{equation}
    \mathcal{L}_r = \mathbb{E}\left[\Vert (\textbf{e}_i^s - \textbf{e}_i^{ {s \to t} })-( \textbf{t}_i^s-\textbf{t}_i^t ) \Vert_1\right].
\end{equation}
The idea behind RDD is to guide the model in directly searching the generic embedding.
This is achieved by refining the prompt representations through adjusting the relative positions of the image representations based on their language representations.

\textbf{Positive-Negative Separation.}
DetPro \cite{du2022learning} learns diverse contextual prompt representations by training on cropped proposals across varying IoU thresholds.
Similarly, we adopt an instance-level training strategy by separating positive and negative proposals to capture diverse contexts of target domains.
Initially, instance prompt representations $\textbf{t}_p^t(c)$ and $\textbf{t}_p^t(\textsmaller{\mathcal{C}_{bg}})$ are generated by $\mathcal{T}$ from $\mathcal{R}_p^t(c)$ and $\mathcal{R}_p^t(\textsmaller{\mathcal{C}_{bg}})$.
The representations of positive and negative proposals $\textbf{e}_p^{s \to t}(c)$ and $\textbf{e}_p^{s \to t}(\textsmaller{\mathcal{C}_{bg}})$ are then derived by ROI-Align.
For the positive proposal, the probability belonging to a foreground category $c \in C $ is computed as:
\begin{equation}
     P_{pc} = \frac{\exp(f( \textbf{e}_p^{s \to t}(c),\textbf{t}_p^t(c)))}{\sum\limits_{c' \in \mathcal{C}} \exp(f( \textbf{e}_p^{s \to t}(c),\textbf{t}_p^t(c') ) ) }.
     \label{eq.pc}
\end{equation}
Subsequently, the positive proposal loss is defined as:
\begin{equation}
    \mathcal{L}_{c} = \mathbb{E}\big[-\sum\limits_{c \in \mathcal{C}} y_c \log P_{pc}\big].
\end{equation}
For the negative proposal, the probability belonging to background category is computed as:
\begin{equation}
     P_{nc} = \frac{\exp(f( \textbf{e}_p^{s \to t}(\scalebox{0.8}{$\mathcal{C}_{bg}$}),\textbf{t}_p^t(\scalebox{0.8}{$\mathcal{C}_{bg}$}) ))}{\sum\limits_{c' \in \mathcal{C} \cup  \scalebox{0.7}{$\mathcal{C}_{bg}$} } \exp(f( \textbf{e}_p^{s \to t}(\scalebox{0.8}{$\mathcal{C}_{bg}$}),\textbf{t}_p^t(c') ) ) }.
     \label{eq.nc}
\end{equation}
Then, the loss of negative proposal is defined as:
\begin{equation}
    \mathcal{L}_{\textsmaller{{bg}}} = \mathbb{E}\big[-\sum_{ c \in \mathcal{C} \cup  \textsmaller{\mathcal{C}_{bg}} } y_{\textsmaller{{bg}}} \log P_{nc}\big],
\end{equation}
where $y_{\textsmaller{{bg}}}$ is the reciprocal of the total number of classes, including the background class. By computing the difference from $y_{\textsmaller{{bg}}}$ and only considering positive differences, we limit the overconfidence of the background class.

\subsection{Zero-shot Training and Testing \label{sec.3.4}}
The training process consists of two stages, both conducted on the source domain.
In the first stage, as illustrated in Fig.~\ref{fig:3}, we perform prompt and enhancement learning. 
Initializing MDP parameters randomly, we train URE to tranform source features into stylized pseudo-target representations.
Simultaneously, these synthetic features drive MDP adaptation through alignment between visual and language representations.
These processes are performed concurrently: the updated MDP refines URE, and the improved URE further enhances the pseudo target domain features, which in turn refine the MDP.

In the second stage, we freeze the optimized MDP, URE and text encoder $\mathcal{T}$ to preserve acquired domain knowledge, then fine-tune the CLIP backbone and RCNN detector.
Following CLIP-GAP~\cite{vidit2023clip}, we randomly apply enhancements to source features with a probability of 50\%.
The bounding boxes are predicted with a regression head, while classification scores of proposals are computed via CLIP.
The detector is trained with both regression and classification losses, following the framework of Faster R-CNN \cite{ren2015faster}.

During inference on target domains, we disable URE transformations while maintaining identical regression and classification procedures, ensuring consistent feature processing between training and deployment phases.

\begin{figure}[t]
  \centering
  \setlength{\belowcaptionskip}{-0.2cm}
   \includegraphics[width=0.90\linewidth]{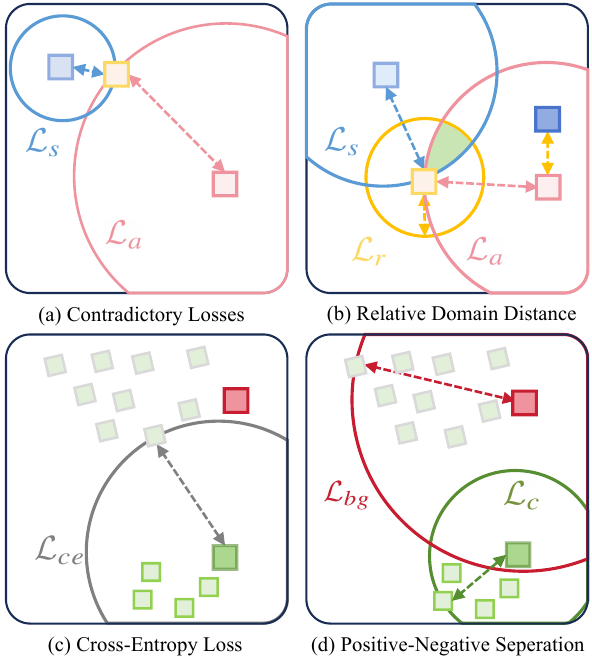}

   \caption{
\textbf{Illustration of Our Proposed Strategies.} 
Contradictory losses represent $\mathcal{L}_a$ and $\mathcal{L}_s$. 
The symbols 
\makebox[5pt][l]{\includegraphics[height=1.0em]{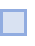}}~and
\makebox[5pt][l]{\includegraphics[height=1.0em]{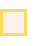}} denote the representations of the source domain, pseudo target domain.
\makebox[5pt][l]{\includegraphics[height=1.0em]{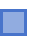}}~and~ 
\makebox[5pt][l]{\includegraphics[height=1.0em]{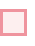}} represent the representations of image prompts for the source and target domains. 
Additionally, the cross-entropy loss $\mathcal{L}_{ce}$ assigns proposals to all possible foregrounds. 
The symbols 
\makebox[5pt][l]{\includegraphics[height=1.0em]{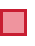}}, 
\makebox[5pt][l]{\includegraphics[height=1.0em]{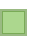}}, 
\makebox[5pt][l]{\includegraphics[height=1.0em]{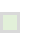}}, and 
\makebox[5pt][l]{\includegraphics[height=1.0em]{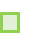}} 
indicate the representations of negative prompt, positive prompt, background objects, and foreground objects, respectively.
}
   \label{fig:5}
\end{figure}

\subsection{Discussion}
The transformation of source domain features requires a careful balance between preserving semantic integrity and enhancing feature style diversity. 
To achieve this, we need to align multi-modal representations while also address challenges arising from latent space variability and background contextual interference.
Specifically, excessive enhancement must be controlled to avoid distorting the critical semantic information of the objects.
However, the objectives of enhancement and constraint are inherently contradictory.
As illustrated in Fig.~\ref{fig:5}(a), these two objectives interfere with each other during training, causing fluctuations that hinder the efficient learning of a generalized representation.
To address this issue, we introduce representation regularization loss $\mathcal{L}_r$ to explore the multi-modal representation relationships between the source and target domains.
It can constrain the search within the green overlap region, thereby stabilizing the prompt and enhanement learning at image level, as illustrated in Fig.~\ref{fig:5}(b).

Proposals in object detection are categorized as positive or negative based on their IoU scores.
These proposals, however, often contain not only the target object but also contextual information from the background and surrounding objects.
As illustrated in Fig.~\ref{fig:5}(c), the vanilla cross-entropy loss is insufficient to model the varying latent space, thereby undermining the adaptation performance.
To this end, we introduce $\mathcal{L}_c$ to shrink search space for foreground object and $\mathcal{L}_{\textsmaller{bg}}$ to assist former detecting background classes as shown in Fig.~\ref{fig:5}(d). 
Different from DetPro\cite{du2022learning}, which relies on pre-cropped proposals and only updates shared prompt representations with category keywords, our method cooperatively integrates prompt optimization with cross-domain representation synthesis.
Most importantly, both the multi-view prompt representation and the enhancement representation are jointly trained in a unified framework.

\section{Experiments}
\subsection{Datasets}
We adopt nine benchmark object detection datasets across three cross-domain settings.

\textbf{Diverse Weather Conditions.}
This setting incorporates five datasets representing various weather conditions from BDD100K \cite{yu2020bdd100k}, FoggyCityscapes \cite{sakaridis2018semantic}, and Adverse-Weather \cite{hassaballah2020vehicle}.
Daytime Clear dataset comprises 19,395 training images.
Night Clear dataset includes 26,158 images.
Dusk Rainy and Night-Rainy datasets contain 3,501 and 2,494 images, respectively.
Daytime Foggy dataset consists of 3,775 images.
Training is exclusively conducted using the Daytime Clear dataset, while testing is performed on other adverse weather datasets.

\begin{figure*}
  \centering
  \setlength{\abovecaptionskip}{0.2cm}
  \setlength{\belowcaptionskip}{-0.5cm}
   \includegraphics[width=0.95\linewidth]{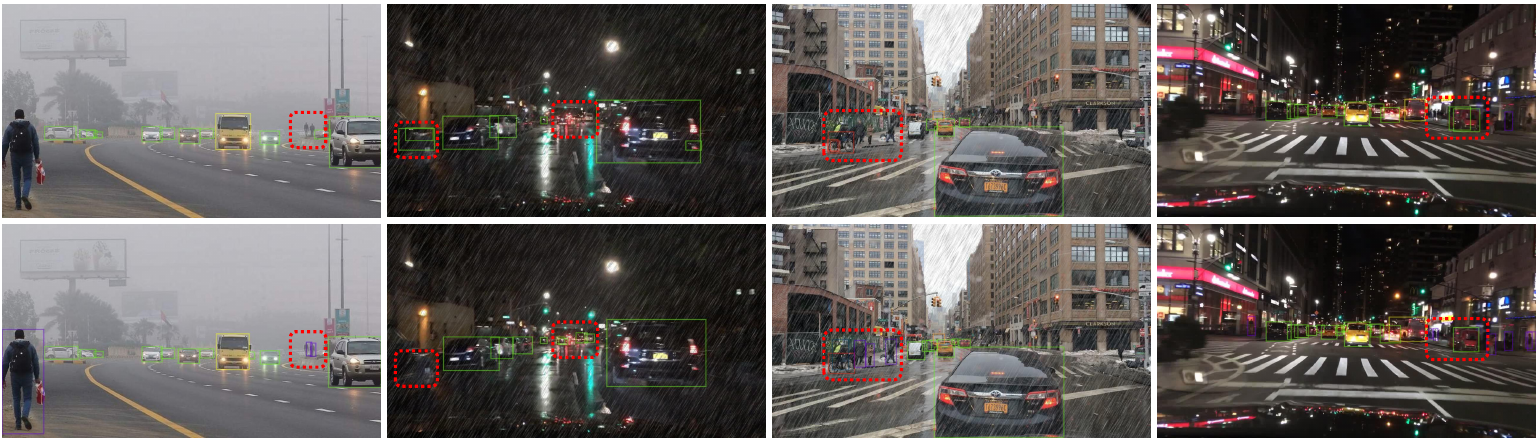}
   \caption{\textbf{Qualitative Results on Diverse Weather Conditions}.
   The top and bottom rows donate the results of OA-DG \cite{lee2024object} and our proposed UPRE, respectively.
   }
   % \vspace{-2mm}
   \label{fig:4}
\end{figure*}

\textbf{Cross-City Scenarios.}
This setting involves datasets collected from different cities, highlighting geographical domain shifts.
Specifically, Cityscapes dataset \cite{cordts2016cityscapes}, comprising 2,975 images, is gathered from Germany.
BDD100K dataset, collected across the US, contains 100k images, where 47,060 images captured under clear daytime are utilized.
KITTI dataset \cite{geiger2012we} from Germany provides 7,481 labeled images.
While Cityscapes and BDD100K share the same seven categories, we exclusively report results for the \textit{car} category in KITTI.
In this scenario, Cityscapes serves as the source domain, and BDD100K and KITTI serve as the target domains.

\textbf{Virtual-to-Real World Transitions.}
Sim10K dataset \cite{johnson2016driving}, consisting of 10k synthetic images rendered from the Grand Theft Auto V game, is used as the source domain.
The target domains in this setting are Cityscapes, BDD100K, and KITTI, and the evaluations are focused on the \textit{car} category.

\subsection{Implementation}
We employ Faster R-CNN \cite{ren2015faster} with backbone initialized from the pre-trained CLIP ResNet-101 \cite{he2016deep}.
During training, input images are resized to 600$\times$1067.
The text encoder and the first two blocks of the image encoder remain frozen across all phases.
Our model is optimized using SGD with momentum of 0.1, a weight decay of 10$^{-4}$.
All experiments are conducted on the Detectron2 platform using an NVIDIA 4090 GPU with a batch size of 4.

\textbf{Prompt and enhancement Learning.}
Initially, we perform a warm-up, training the detector for 1.3k iterations on the source dataset with a learning rate of 0.001.
We then proceed to prompt and enhancement training, where source features from the first layer are integrated with enhancements.
This phase spans 5k iterations with a learning rate of 0.001, reduced by a factor of 10 at iteration 4.5k.

\textbf{Detector Fine-tuning.}
During the fine-tuning phase, we freeze the parameters of MDP and URE.
Training is conducted for 100k iterations, starting with a learning rate of 0.001, which decays by a factor of 0.1 at 40k iterations.

\begin{table}[t]
  \centering
  \setlength{\abovecaptionskip}{0.2cm}
  \setlength{\belowcaptionskip}{-0.5cm}
  \setlength{\heavyrulewidth}{1.5pt}
  \belowrulesep=0pt
  \aboverulesep=0pt
  \resizebox{8.3cm}{!}{
  \begin{tabular}{l|c|cccc}
    \toprule 
    \multirow{2}{*}{Method} & \multirow{2}{*}{Venues}  & Daytime & Night & Night & Dusk \\
    &  & Foggy & Clear & Rainy & Rainy \\
    \midrule
    Faster-RCNN\cite{ren2015faster} & NeurIPS'15  & 32.0 & 34.4 & 12.4 & 26.0 \\
    S-DGOD \cite{wu2022single} & CVPR'22  & 33.5 & 36.6 & 16.6 & 28.2 \\
    CLIP-GAP \cite{vidit2023clip} & CVPR'23  & 38.5 & 36.9 & 18.7 & 32.3 \\
    PODA* \cite{fahes2023poda} & ICCV'23 & 39.2 & 38.7 & 19.0 & 33.4 \\
    OA-DG \cite{lee2024object} & AAAI'24  & 38.3 & 38.0 & 16.8 & \underline{33.9} \\
    DAI-Net*\cite{du2024boosting} & CVPR'24  & 36.7 &\underline{41.0} & 18.9 & 33.0 \\
    PDD\cite{li2024prompt} & CVPR'24  & 39.1 & 38.5 & \underline{19.2} & 33.7 \\
    UFR\cite{liu2024unbiased} & CVPR'24  & \underline{39.6} &{40.8} & \underline{19.2} & 33.2 \\
    \midrule \midrule
    UPRE(Ours) & -  & \textbf{40.0} & \textbf{41.5} & \textbf{19.8} & \textbf{34.5} \\
    \bottomrule
  \end{tabular}}
  \caption{Quantitative results on Diverse Weather Conditions. Results marked with (*) indicate those obtained through replication based on Faster-RCNN.}
  \label{tab:Quantitative results on Diverse Weather Conditions}
\end{table}
\subsection{Main Results}
We compare our method with the S-DGOD in object detection methods OA-DG \cite{lee2024object}, PDD \cite{li2024prompt}, FDA \cite{danish2024improving} and UFR \cite{liu2024unbiased}.
Besides, we also compare our method with the ZSDA in object detection methods DAI-Net \cite{du2024boosting} and PODA \cite{fahes2023poda}. 
For each dataset, we follow the same protocol of existing works, and report the average precision (AP$_{50}$) of each class and the mean average precision (mAP) over all classes.

\textbf{Diverse Weather Conditions.}
The qualitative and quantitative evaluation results are presented in Tab.~\ref{tab:Quantitative results on Diverse Weather Conditions} and Fig.~\ref{fig:4}.
On average, our method improves mAP by 7.8\% over Faster R-CNN across all conditions.
In the Daytime Foggy scenario (Tab.~\ref{tab:Per-class results on Daytime Clear to Daytime Foggy}), UPRE achieves an impressive 40.0\% mAP, surpassing Faster R-CNN by 8.0\%. Notably, we observe significant gains for challenging classes such as bus (+6.8\%), motor (+9.9\%), and rider (+9.1\%). These improvements highlight our method's ability to capture fine-grained contextual details under low-visibility foggy conditions.

For the Night Rainy condition (Tab.~\ref{tab:Per-class results on Daytime Clear to Night Rainy}), UPRE achieves a top mAP of 19.8\%, surpassing Faster R-CNN by an average margin of 7.4\%. Besides, we observe improvements in detecting occluded objects, such as bike (+4.2\%) and rider (+4.2\%). This demonstrates the effectiveness of our approach in handling nighttime rain effects, where poor illumination and reflective surfaces often degrade detection performance.
In the Dusk Rainy setting (Tab.~\ref{tab:Per-class results on Daytime Clear to Dusk Rainy}), UPRE achieves 34.5\% mAP, surpassing Faster R-CNN and CLIP-GAP by 8.5\% and 2.2\%, respectively. Particularly, we achieve remarkable gains of 13.4\% and 6.4\% mAP for bike, underscoring the adaptability to transitional lighting and rainy conditions.
Finally, in the Night Clear scenario (Tab.~\ref{tab:Per-class results on Daytime Clear to Night Clear}), our method achieves the best performance with an mAP of 41.5\%, further confirming its adaptability across diverse scenarios.

\begin{table}
  \centering
  \setlength{\abovecaptionskip}{0.2cm}
  \setlength{\belowcaptionskip}{-0.3cm}
  \setlength{\heavyrulewidth}{1.5pt}
  \belowrulesep=0pt
  \aboverulesep=0pt
  \resizebox{8.3cm}{!}{
  \begin{tabular}{l|ccccccc|>{\columncolor[gray]{0.9}}c}
    \toprule
    Method & Bus & Bike & Car	& Motor	& Person & Rider & Truck & mAP \\
    \midrule
    Faster-RCNN\cite{ren2015faster} & 30.7 & 26.7 & 49.7 & 26.2 & 30.9 & 35.5 & 23.2 & 32.0\\
    S-DGOD\cite{wu2022single} & 32.9 & 28.0 & 48.8 & 29.8 & 32.5 & 38.2 & 24.1 & 33.5\\
    CLIP-GAP\cite{vidit2023clip} & 36.1 & 34.3 &	58.0 & 33.1 & 39.0 & 43.9 & 25.1 & 38.5\\
    PODA* \cite{fahes2023poda} &34.6 &34.1 &50.7 &31.2 &38.5 &44.0 &25.9 & 39.2  \\
    OA-DG\cite{lee2024object} &- &- &- &- &- &- &- & 38.3  \\
    DAI-Net*\cite{du2024boosting} & 33.7 & 32.8 & 57.9 & 33.1 & 37.9 & 41.0 & 24.7 & 36.7\\
    PDD\cite{li2024prompt}	& 36.1 & 34.5 & \underline{58.4} & 33.3	& \textbf{40.5} & 44.2 &	26.2 & 39.1\\
    UFR\cite{liu2024unbiased}	& \underline{36.9} & \textbf{35.8} &	\textbf{61.7} & \underline{33.7} & \underline{39.5} & 42.2 &	\textbf{27.5} & 39.6\\
    \midrule \midrule
    UPRE(Ours) & \textbf{37.5}	& \underline{35.6} & \underline{58.4} & \textbf{36.1} & 39.0 &  \textbf{44.6} & \underline{27.0} & \textbf{40.0}\\
    \bottomrule
  \end{tabular}
  }
  \caption{Per-class results on Daytime Clear to Daytime Foggy.}
  % \vspace{-2mm}
  \label{tab:Per-class results on Daytime Clear to Daytime Foggy}
\end{table}

\begin{table}
  \centering
  \setlength{\abovecaptionskip}{0.2cm}
  \setlength{\belowcaptionskip}{-0.5cm}
  \setlength{\heavyrulewidth}{1.5pt}
  \belowrulesep=0pt
  \aboverulesep=0pt
  \resizebox{8.3cm}{!}{
  \begin{tabular}{l|ccccccc|>{\columncolor[gray]{0.9}}c}
    \toprule
    Method & Bus & Bike & Car	& Motor	& Person & Rider & Truck & mAP \\
    \midrule
    Faster-RCNN\cite{ren2015faster} & 22.6 & 11.5 &	27.7 & 0.4 & 10	& 10.5 & 19.0 & 12.4 \\
    S-DGOD\cite{wu2022single} & 24.4 & 11.6 & 29.5 & 9.8 & 10.5 & 11.4 & 19.2 & 16.6\\
    CLIP-GAP\cite{vidit2023clip} & 28.6 & 12.1 &	36.1 & 9.2 & 12.3 &	9.6	& 22.9 & 18.7\\
    PODA* \cite{fahes2023poda} &26.8 &11.7 &36.4 &9.0 &12.7 &11.5 &28.8 & 19.0  \\
    OA-DG\cite{lee2024object} & - &- &- &- &- &- &- & 16.8  \\
    DAI-Net*\cite{du2024boosting} & 24.6 & 11.7 & \textbf{37.5} & 9.0 & 12.9 & 11.2 & 21.4 & 18.9\\
    PDD\cite{li2024prompt}	& 25.6 & \underline{12.1} & 35.8 & \textbf{10.1} & \textbf{14.2} & \underline{12.9} & 22.9 & 19.2\\
    UFR\cite{liu2024unbiased}	& \textbf{29.2} & 11.8 & 36.1 & \underline{9.4}	& {13.1} & 10.5 & \underline{23.3} & 19.2\\
    \midrule \midrule
    UPRE(Ours) & \underline{27.7} & \textbf{15.7} & \underline{36.7} & \underline{9.4} & \underline{13.5} & \textbf{14.7} & \textbf{25.7} & \textbf{19.8}\\
    \bottomrule
  \end{tabular}
  }
  \caption{Per-class results on Daytime Clear to Night Rainy.}
  % \vspace{-2mm}
  \label{tab:Per-class results on Daytime Clear to Night Rainy}
\end{table}

\textbf{Cross-City Scenarios.}
Geographic domain shifts between cities pose a significant challenge due to varying urban layouts and scene characteristics. 
However, previous methods \cite{danish2024improving,vidit2023clip,liu2024unbiased,lee2024object,wu2024g} largely overlook cross-city adaptation scenarios. 
To address this limitation, we propose a novel approach that emphasizes more expressive prompt representations for handling cross-city variations.

Cityscapes primarily focuses on street scenes, while BDD100K and KITTI encompass diverse road environments, including highways and parking lots. 
To bridge these differences, we introduce a novel prompt representation $k_d$, which integrates weather and city scene descriptors. 
This representation enables the model to simulate knowledge transfer across different weather conditions and urban scenes, significantly enhancing its ability to handle geographic domain shifts.
Quantitative results for Cityscapes to BDD100K and KITTI are presented in Tab.~\ref{tab:Quantitative results on Cross-City Scenarios}. 
Compared to state-of-the-art methods such as CLIP-GAP, PODA*, OA-DG, and DAI-Net*, our method (UPRE) achieves average mAP improvements of 1.9\%, 1.4\%, 1.4\%, and 4.5\%, respectively.

\begin{table}
  \centering
  \setlength{\abovecaptionskip}{0.2cm}
  \setlength{\belowcaptionskip}{-0.3cm}
  \setlength{\heavyrulewidth}{1.5pt}
  \belowrulesep=0pt
  \aboverulesep=0pt
  \resizebox{8.3cm}{!}{
  \begin{tabular}{l|ccccccc|>{\columncolor[gray]{0.9}}c}
    \toprule
    Method & Bus & Bike & Car	& Motor	& Person & Rider & Truck & mAP \\
    \midrule
    Faster-RCNN\cite{ren2015faster} & 36.8 & 15.8 &	50.1 & 12.8 & 18.9 & 12.4 &	39.5 & 26.0 \\
    S-DGOD\cite{wu2022single} & 37.1 & 16.9 & 50.9 & 13.4 & 19.7 & 16.3 & 40.7 & 28.2\\
    CLIP-GAP\cite{vidit2023clip} & 37.8 & 22.8 & 60.7 & 16.8 & 26.8 & 18.7 &	42.4 & 32.3 \\
    PODA* \cite{fahes2023poda} &38.2 &24.9 &58.1 &18.5 &27.5 &17.7 &43.6 & 33.4  \\
    OA-DG\cite{lee2024object} &- &- &- &- &- &- &- & 33.9  \\
    DAI-Net*\cite{du2024boosting} & 36.9 & \underline{26.1} & 59.0 & 17.5 & 28.2 & 16.5 & 42.7 & 33.0  \\
    PDD\cite{li2024prompt} & \textbf{39.4} &	25.2&	60.9 &	\textbf{20.4}	& \textbf{29.9}	&16.5	& \underline{43.9} &	33.7 \\
    UFR\cite{liu2024unbiased} &	37.1	& 21.8 & \textbf{67.9} &	16.4&	27.4&	\underline{17.9} &	\underline{43.9} &	33.2 \\
    \midrule \midrule
    UPRE(Ours) & \underline{39.1}	& \textbf{29.2} & \underline{65.5} &	\underline{19.7} &	\underline{28.6} &\textbf{19.2} &	\textbf{45.7} &	\textbf{34.5} \\
    \bottomrule
  \end{tabular}
  }
  \caption{Per-class results on Daytime Clear to Dusk Rainy.}
   % \vspace{-2mm}
  \label{tab:Per-class results on Daytime Clear to Dusk Rainy}
\end{table}

\begin{table}
  \centering
  \setlength{\abovecaptionskip}{0.2cm}
  \setlength{\belowcaptionskip}{-0.5cm}
  \setlength{\heavyrulewidth}{1.5pt}
  \belowrulesep=0pt
  \aboverulesep=0pt
  \resizebox{8.3cm}{!}{
  \begin{tabular}{l|ccccccc|>{\columncolor[gray]{0.9}}c}
    \toprule
    Method & Bus & Bike & Car	& Motor	& Person & Rider & Truck & mAP \\
    \midrule
    Faster-RCNN\cite{ren2015faster} & 37.7 & 30.6 & 49.5 & 15.4 & 31.5 & 28.6 & 40.8 & 34.4\\
    S-DGOD\cite{wu2022single} & 40.6 & 35.1 & 50.7 & 19.7 & 34.7 & 32.1 & 43.4 & 36.6\\
    CLIP-GAP\cite{vidit2023clip} & 37.8 & 22.8 & 60.7 & 16.8 & 26.8 & 18.7 &	42.4 & 36.9\\
    PODA* \cite{fahes2023poda} &39.6 &34.7 &58.5 &19.5 &37.4 &27.2 &42.6 & 38.7  \\
    OA-DG\cite{lee2024object} &- &- &- &- &- &- &- & 38.0  \\
    DAI-Net*\cite{du2024boosting} & 41.6 & \underline{38.6} & 59.2 & 20.5 & 45.2 & \underline{31.1} & 43.7 & \underline{41.0}\\
    PDD\cite{li2024prompt}	& 40.9 & 35.0 &	59.0 & \textbf{21.3} & 40.4 & 29.9  & 42.9 &	38.5\\
    UFR\cite{liu2024unbiased}	& \textbf{43.6} & 38.1 & \textbf{66.1} & 14.7 & \textbf{49.1} & 26.4 &	\textbf{47.5} & {40.8}\\
    \midrule \midrule
    UPRE(Ours) & \underline{42.4}	& \textbf{38.7} & \underline{61.4} &	\underline{20.6} & \underline{45.7} & \textbf{31.6} & \underline{44.3} & \textbf{41.5}\\
    \bottomrule
  \end{tabular}
  }
  \caption{Per-class results on Daytime Clear to Night Clear.}
  % \vspace{-2mm}
  \label{tab:Per-class results on Daytime Clear to Night Clear}
\end{table}

 \textbf{Virtual to Real World Transitions.}
The virtual world of GTAV completely differs from the real world in style. 
To address this gap, we harness easily accessible virtual data to facilitate detector adaptation to unseen real-world domains.
Specifically, we integrate real-world weather conditions, such as rain, fog, and night, into the sunny virtual world, transforming it into a more realistic and diverse representation. 
This approach ensures consistent and transferable prompt representations across domains under varying weather conditions.

By simulating real world weather effects in the virtual domain, our method bridges the gap between synthetic and real world data.  
Quantitative results for virtual-to-real world transitions are presented in Tab.~\ref{tab:Results on Virtual-to-Real World Transitions}.
Our method outperforms existing approaches across three real-world datasets, achieving mAP improvements of 1.6\%, 1.5\%, 1.8\%, and 7.3\% compared to CLIP-GAP, PODA*, OA-DG, and DAI-Net*, respectively.  
The superior performance of our method can be attributed to its ability to effectively simulate real-world conditions within the virtual domain.

\begin{figure}[t]
\setlength{\abovecaptionskip}{0.1cm}
\setlength{\belowcaptionskip}{-0.1cm}
\begin{center}
\begin{tabular}{c@{}c}
\hspace{-3mm}
\includegraphics[width=0.5\linewidth,height=2.7cm]{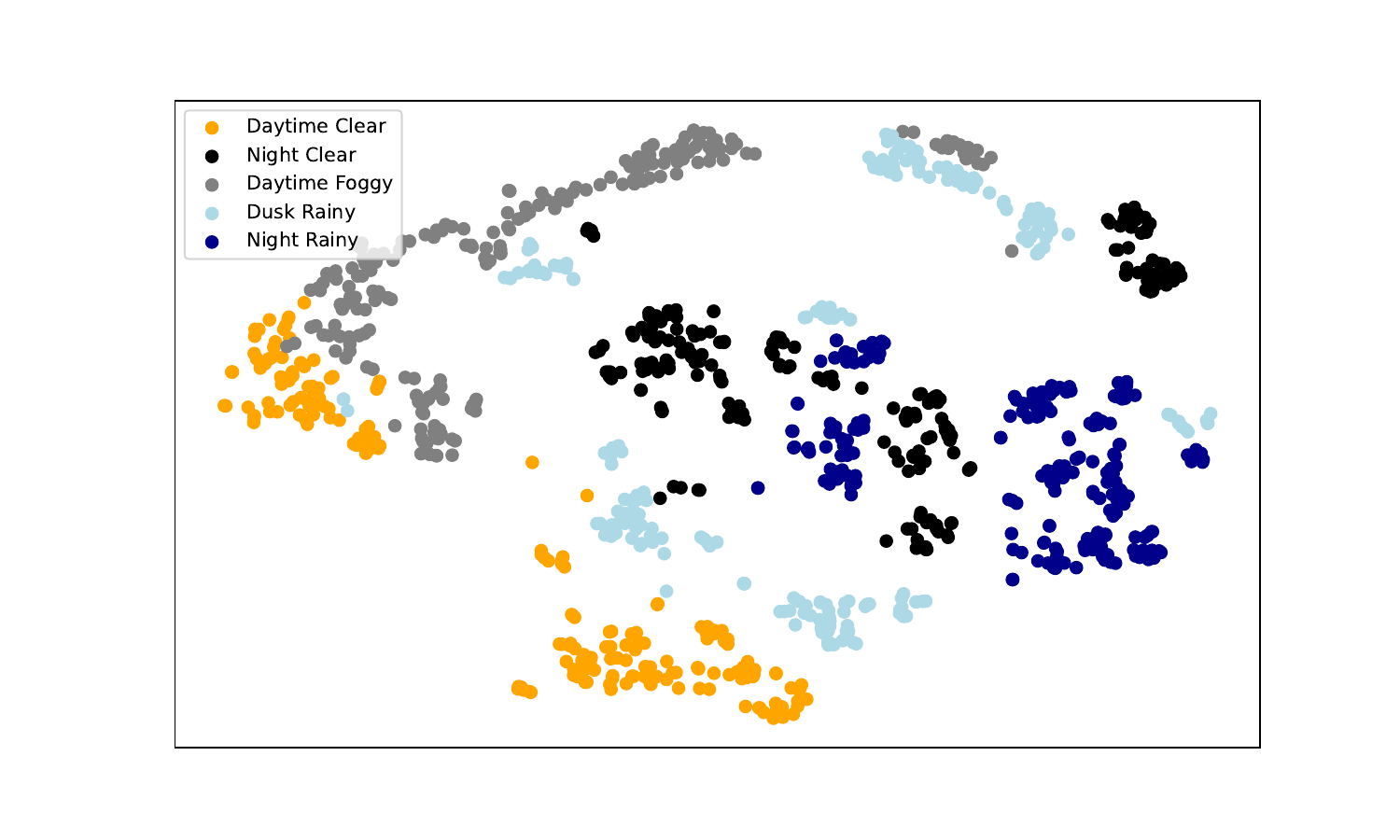} \ &
\includegraphics[width=0.5\linewidth,height=2.7cm]{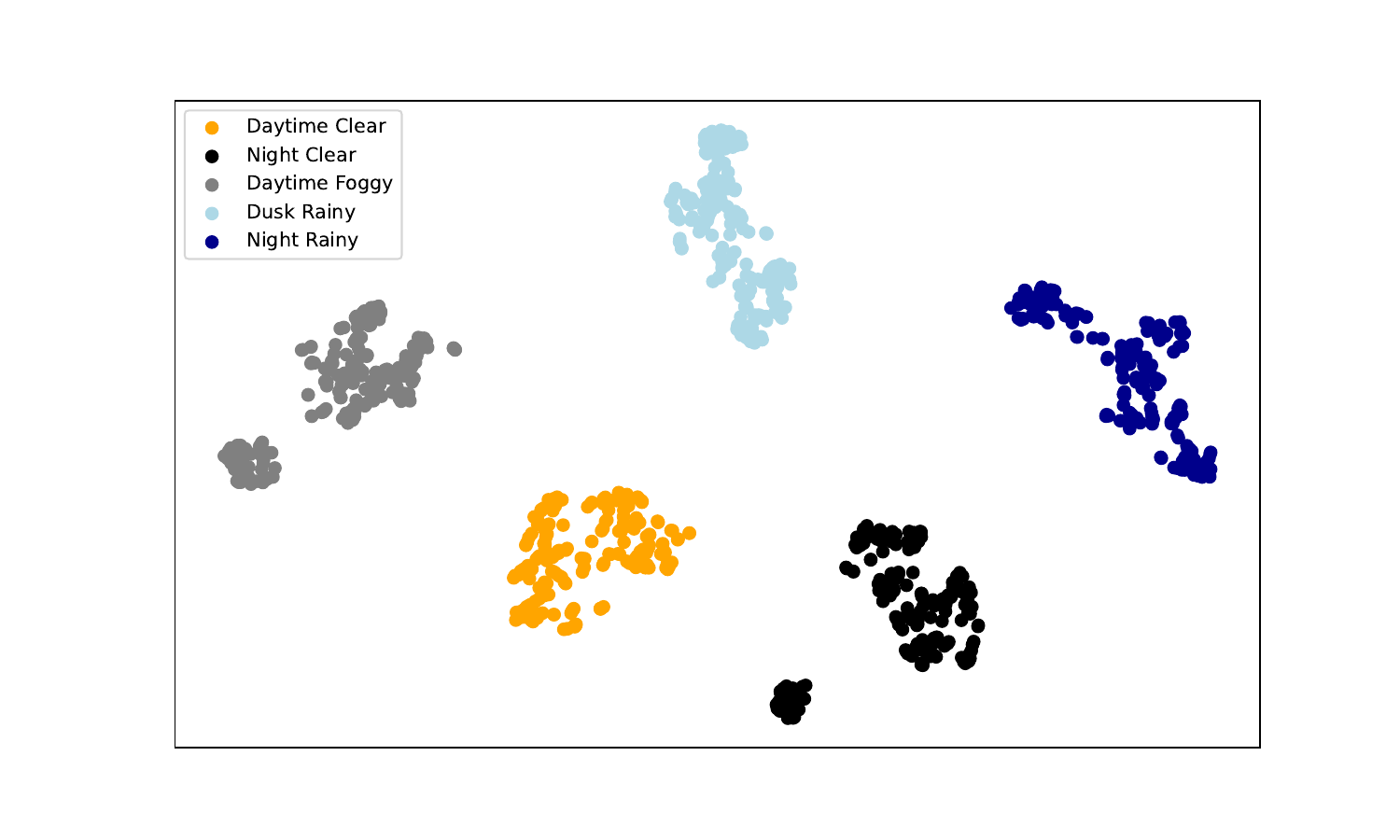} \\
\end{tabular}
\end{center}
\vspace{-5mm}
\caption{The t-SNE visualization of image embeddings across five domains using identical images and experimental settings. CLIP (Left) represents coarse generalization capabilities, whereas UPRE (Right) exhibits superior adaptation to each target domain.}
% \vspace{-5mm}
\label{t-SNE}
\end{figure}

   \textbf{The visualization of the domain embeddings.}
As depicted in Fig.~\ref{t-SNE} (left), although CLIP possesses generalization abilities, it lacks discrimination in embeddings for different weather conditions, such as the entanglement of Night Clear and Night Rainy embeddings. In Fig.~\ref{t-SNE} (right), UPRE enhances CLIP’s performance and achieves superior domain adaptation capability.

\subsection{Ablation Study}
\textbf{Effect of Prompt Setting.}
Tab.~\ref{tab:Effect study of prompt Setting} compares different prompt settings, where either complete descriptions or only keywords are utilized.
The learnable prompt with a complete prompt structure improves mAP by an average of 3.0\%, demonstrating its effectiveness in capturing cross-domain knowledge.  
By maintaining the complete prompt, the learnable context can focus on acquiring domain-specific knowledge critical for cross-domain object detection, as described in Sec.~\ref{sec.3.2}.  
However, we find that using a learnable context with shared parameters hurts performance, e.g., reducing mAP by 2.4\% on Night Rainy.
This underscores the importance of prompt design, as mere learnability without proper constraints or separation is suboptimal. 
\begin{table}
  \centering
  \setlength{\abovecaptionskip}{0.2cm}
  \setlength{\belowcaptionskip}{-0.1cm}
  \setlength{\heavyrulewidth}{1.5pt}
  \belowrulesep=0pt
  \aboverulesep=0pt
  \resizebox{8.3cm}{!}{
  \begin{tabular}{l|ccccccc>{\columncolor[gray]{0.9}}c|c}
    \toprule
    \multicolumn{1}{c}{\multirow{2}{*}{Method}} & \multicolumn{8}{c}{BDD100K} & KITTI \\
    \cline{2-10}
     \multicolumn{1}{c}{} & Bus & Bike & Car	& Motor	& Person & Rider & Truck
 & \multicolumn{1}{c}{mAP} & AP of Car\\
    \midrule
    Faster-RCNN\cite{ren2015faster} & 21.9 & 22.7 &	36.9 & 17.8  & 24.1 & 25.4 & 22.6 & 24.5 & \cellcolor[gray]{0.9}72.5 \\
    CLIP-GAP\cite{vidit2023clip} & 22.4 & 23.2 &	\underline{47.2} & \underline{20.2} & 20.1 & 30.2 &	15.0 & 26.3 & \cellcolor[gray]{0.9}72.9 \\
    PODA* \cite{fahes2023poda} & \underline{23.1} & 22.7 &	{46.5} & \textbf{20.4} & 20.9 & \underline{31.0} &	18.5 & 26.7 & \cellcolor[gray]{0.9}\underline{73.6} \\
    OA-DG\cite{lee2024object} & 17.3 & \underline{24.0} & 42.8 & 19.6 & \underline{33.7} & 28.2 & \underline{20.3} & \underline{27.2} & \cellcolor[gray]{0.9}73.0 \\
    DAI-Net*\cite{du2024boosting} & 18.1 & 22.0 & 36.9 & 17.3 & 30.1 & 29.2 & 18.4 & 22.8 & \cellcolor[gray]{0.9}71.3\\
    \midrule \midrule
    UPRE(Ours) & \textbf{23.8} &	\textbf{24.5} & \textbf{48.7} & 19.7 & \textbf{34.7} &	\textbf{31.9} & \textbf{21.5} & \textbf{28.7} & \cellcolor[gray]{0.9}\textbf{74.3} \\
    \bottomrule
  \end{tabular}
  }
  \caption{Quantitative results on Cross-City Scenarios.}
  \vspace{-4mm}
  \label{tab:Quantitative results on Cross-City Scenarios}
\end{table}

\begin{table}
  \centering
  \setlength{\abovecaptionskip}{0.2cm}
  \setlength{\belowcaptionskip}{-0.5cm}
  \setlength{\heavyrulewidth}{1.5pt}
  \setlength{\tabcolsep}{12pt}
  \belowrulesep=0pt
  \aboverulesep=0pt
  \resizebox{6.3cm}{!}{
  \begin{tabular}{l|c|c|c}
    \toprule
    Method & Cityscape & BDD100K & KITTI\\
    \midrule
    Faster-RCNN\cite{ren2015faster}	& 34.3 & 29.8 &	47.0 \\
    CLIP-GAP\cite{vidit2023clip} & 46.8 & 35.3 &	\underline{60.7} \\
    PODA* \cite{fahes2023poda} & 46.5 & \underline{36.1} & 60.3 \\
    OA-DG\cite{lee2024object} & \underline{47.0} & {35.6} & 59.5 \\
    DAI-Net*\cite{du2024boosting} & 41.8 & 30.7 & 53.1 \\
    \midrule \midrule
    UPRE(Ours) & \textbf{47.9} & \textbf{37.8} & \textbf{61.9} \\
    \bottomrule
  \end{tabular}
  }
  \caption{Quantitative results on Virtual-to-Real World Transitions.}
  % \vspace{-2mm}
  \label{tab:Results on Virtual-to-Real World Transitions}
\end{table}

\textbf{Analysis of Image-level Enhancement.}
We study various loss combinations in the RDD in Tab.~\ref{tab:Analysis of the Relative Domain Distance Enhance}.
Our default setting, which includes $\mathcal{L}_a$, $\mathcal{L}_s$ and $\mathcal{L}_r$, achieves the best and most stable performance compared to other configurations. 
Using only $\mathcal{L}_a$ leads to performance degradation, while adding $\mathcal{L}_s$ can result in instability due to conflicting optimization objectives. 
Notably, incorporating $\mathcal{L}_r$ into the enhancement further averagely improves mAP by 0.8\% and MAD by 1.3\%, demonstrating its effectiveness in aligning multi-modal representations and stabilizing training. 
This highlights the importance of balancing enhancement and constraint losses to achieve robust domain adaptation.

%  \textbf{Influence of Instance-level Enhancement.}
% Tab.~\ref{tab:Influence study of the PNS} highlights the effectiveness of our PNS strategy in improving performance.  
% %
% By separating positive and negative proposals and computing foreground ($\mathcal{L}_c$) and background ($\mathcal{L}_{\textsmaller{bg}}$) losses independently, our method achieves a substantial mAP gain, e.g., +1.6\% on Night Clear.  
% %
% Notably, $\mathcal{L}_{\textsmaller{bg}}$ performs competitively, reducing mAP by average 1.0\% compared to the vanilla cross-entropy loss ($\mathcal{L}_{ce}$).  
% %
% Moreover, $\mathcal{L}_c$ alone achieves comparable performance to the combined loss ($\mathcal{L}_{\textsmaller{bg}} + \mathcal{L}_c$), underscoring the importance of effectively modeling background context.
%

\begin{table}
  \centering
  \setlength{\abovecaptionskip}{0.2cm}
  \setlength{\belowcaptionskip}{-0.1cm}
  \setlength{\heavyrulewidth}{1.5pt}
  \belowrulesep=0pt
  \aboverulesep=0pt
  \resizebox{8.5cm}{!}{
  \begin{tabular}{c|c|cccc}
    \toprule
    Learnable & Static & Daytime Foggy& Night Clear & Night Rainy & Dusk Rainy\\
    \midrule
    w/o  & K & 37.2 & 37.5 & 17.0 & 31.7 \\
    w/o  & C & 38.2 & 39.3 & 17.9 & 32.2 \\
    w  & K & \underline{38.7} & \underline{40.1} & \underline{18.6} & \underline{33.0} \\
    w + S & C & 38.0 & 39.7 & 17.4 & 32.2 \\
    w  & C & \textbf{40.0} & \textbf{41.5} & \textbf{19.8} & \textbf{34.5} \\
    \bottomrule
  \end{tabular}
  }
  \caption{Effect study of prompt setting. w/o and w denote without and with learnable prompt, respectively. S represents using one shared learnable parameters.
  K defines that prompt only contains key word, while C represents using complete prompt in Sec.~\ref{sec.3.2}. }
  % \vspace{-2mm}
  \label{tab:Effect study of prompt Setting}
\end{table}

\begin{table}
  \centering
  \setlength{\abovecaptionskip}{0.2cm}
  \setlength{\belowcaptionskip}{-0.1cm}
  \setlength{\heavyrulewidth}{1.5pt}
  \belowrulesep=0pt
  \aboverulesep=0pt
  \resizebox{8.5cm}{!}{
  \begin{tabular}{c|c|c|cccc|c}
    % \toprule
    % \multicolumn{3}{c}{Method} & \multicolumn{5}{c}{mAP} \\
    \toprule
    $\mathcal{L}_a$ & $\mathcal{L}_s$ & $\mathcal{L}_r$
     & Daytime Foggy & Night Clear & Night Rainy & Dusk Rainy & MAD \\
    \midrule
    \checkmark & - & - & 35.7 & 33.9 & 16.5 & 29.7 & 0.8  \\
    \checkmark & \checkmark & - & \underline{38.0} & \underline{38.0} & \underline{17.7} & \underline{31.3} & 1.7  \\
    - & - & \checkmark & 37.2 & 37.5 & 17.2 & 30.8 & \textbf{0.3}  \\
    \checkmark & \checkmark & \checkmark & \textbf{38.7} & \textbf{38.8} & \textbf{18.3} & \textbf{32.5}  & \underline{0.4}\\
    \bottomrule
  \end{tabular}
  }
  \caption{Analysis of the RDD. The check mark denotes the addition of the loss. MAD is Mean Absolute Deviation, representing discrete fluctuation range of data.}
  \vspace{-5mm}
  \label{tab:Analysis of the Relative Domain Distance Enhance}
\end{table}
%

%
% \begin{table}
%   \centering
%   \setlength{\abovecaptionskip}{0.2cm}
%   \setlength{\belowcaptionskip}{-0.5cm}
%   \setlength{\heavyrulewidth}{1.5pt}
%   \belowrulesep=0pt
%   \aboverulesep=0pt
%   \resizebox{8.5cm}{!}{
%   \begin{tabular}{c|cccc}
%     % \toprule
%     % \multicolumn{3}{c}{Method} & \multicolumn{5}{c}{mAP} \\
%     \toprule
%     Methods & Daytime Foggy & Night Clear & Night Rainy & Dusk Rainy\\
%     \midrule
%     $\mathcal{L}_{ce}$ & 37.0 & 37.3 & 16.5 & 30.8 \\
%     $\mathcal{L}_{ \textsmaller{{bg}}} $ only & 36.6 & 35.7 & 15.5 & 29.6 \\
%     $\mathcal{L}_c$ only & \underline{37.6} & \underline{37.8} & \underline{17.1} & \underline{31.4} \\
%     $\mathcal{L}_{ \textsmaller{{bg}}} $ + $\mathcal{L}_c$ & \textbf{38.7} & \textbf{38.9} & \textbf{17.8} & \textbf{32.8} \\
%     \bottomrule
%   \end{tabular}
%   }
%   \caption{Influence study of the PNS. $\mathcal{L}_{ce}$ is cross-entropy loss.}
%   % \vspace{-2mm}
%   \label{tab:Influence study of the PNS}
% \end{table}

% \vspace{-1mm}
\section{Conclusion}
In this paper, we propose a unified prompt and representation enhancement framework to mitigates both detection and domain biases for ZSDA in object detection.
By jointly optimizing textual prompts and visual enhancements within object detection framework, our method effectively mitigates detection and domain biases.
Specifically, MDP provides language priors for the target domain while also capturing diverse adaptation knowledge for cross-domain object detection.
Meanwhile, URE diversifies domain styles under the guidance of the prompt representations.
Additionally, our multi-level training framework, which incorporates the RDD and PNS enhancement strategies, effectively unlocks the potential of VLMs.
Extensive experiments across multiple settings demonstrate the effectiveness of our approach in breaking the biases mitigation trade-off.

%%%%%%%%% REFERENCES
{\small
\bibliographystyle{ieee_fullname}
\bibliography{egbib}
}
% \newpage
% \maketitlesupplementary
\clearpage
\setcounter{page}{1}
\setcounter{section}{0}
% \maketitlesupplementary
\renewcommand\thesection{\Alph{section}}
\setlist[itemize]{leftmargin=1.6em, label=\textbullet, itemsep=0.1em}

\section*{Appendix}
This supplementary material is organized as follows.

\begin{itemize}
    \item In Section \ref{Additional Experiments}, we provide additional ablation studies to further analyze the effectiveness of the proposed components in UPRE. In particular, we conduct a comprehensive analysis of each proposed modules, test the performance of various enhancement strategies, evaluate the effectiveness of different prompt priors, and examine the impact of training strategies.
    \item In Section \ref{Prompt Engineering}, we list the prompt templates used for zero-shot domain adaptation in object detection.
    \item In Section \ref{More Training Details}, we describe more implementation details of UPRE.
    \item In Section \ref{More Visualization}, we provide additional qualitative visualization results under cross-city scenarios and virtual-to-real world transitions. 
\end{itemize}

\section{Additional Ablation Studies \label{Additional Experiments}}
\quad In this section, we provide additional quantitative experiments to further analyze the effectiveness of components in UPRE.

\noindent \textbf{The Effect of Each Proposed Components.}
To comprehensively evaluate the effectiveness of our proposed method, we conducted extensive ablation studies across its key components. 
The ablation study results are systematically organized in Tab.~\ref{tab:Ablation study of internal modules},
% The detailed results are presented in Tab.~\ref{tab:Ablation study of internal modules}, 
where $Prompt$ denotes the proposed domain adaptation prompt, $Enhance$ represents unified representation enhancement, $Img$ indicates relative domain distance strategy, and $Ins$ signifies positive-negative separation strategy.
First, we investigate the limitations of addressing only one aspect bias. Rows 1-2 reveal that focusing solely on either detection bias ($Prompt$) or domain bias ($Enhance$) leads to suboptimal performance. For instance, Row 1 achieves an mAP of 37.8 on the Daytime Foggy scenario, while Row 2 achieves 38.5. These results underscore the necessity of jointly addressing both biases for effective adaptation.

Next, we validate the efficacy of our proposed RDD (relative domain distance) and PND (positive-negative separation) strategies. Rows 3-4 confirm our theoretical analysis, showing significant improvements when these strategies are applied. Specifically, Row 3 achieves an mAP of 39.2 on Daytime Foggy, compared to Row 4's 38.9, demonstrating the complementary benefits of $Img$ and $Ins$ in enhancing detection performance.
We further analyze the impact of static prompts and image-level alignment. Row 6 simulates the approach used in \cite{vidit2023clip}, revealing that image-level methods fail to effectively fine-tune CLIP, resulting in degraded performance (e.g., mAP of 35.6 on Daytime Foggy). Similarly, Row 7 highlights the importance of instance-level contextual knowledge, showing that learnable prompts are essential as detection knowledge priors for VLMs (mAP of 35.9).

To approximate the effect of proposal-based training in DetPro \cite{du2022learning}, we remove $Img$ and $Enhance$ to train learnable prompts, as shown in Row 10. The results indicate that prompt learning alone is insufficient to overcome domain bias, which achieves an mAP of only 32.1 on Daytime Foggy.
% In contrast, Row 5 demonstrates that adding trainable prompts to DetPro improves mAP by an average of 3.1 across all scenarios, highlighting the effectiveness of our enhancement approach.
In contrast, Row 5 demonstrates that adding trainable prompts to DetPro achieves a consistent mAP improvement of +3.1 across all test scenarios, highlighting the effectiveness of our enhancement approach.

Finally, Rows 8-9 focus on learning image- and instance-level knowledge but fail to handle cross-domain knowledge effectively, achieving mAP values of 32.9 and 32.8, respectively. This limitation underscores the need for our proposed method's comprehensive design, which integrates multiple components to achieve robust zero-shot domain adaptation.
The overall effectiveness of our method is demonstrated in Row 11, where all components ($Prompt$, $Enhance$, $Img$, $Ins$) are combined. These results validate the synergy of our proposed components and their ability to address both detection and domain biases effectively.

\begin{table}
  \centering
  \setlength{\belowcaptionskip}{-0.3cm}
  \setlength{\heavyrulewidth}{1.5pt}
  \belowrulesep=0pt
  \aboverulesep=0pt
  \resizebox{8.3cm}{!}{
  \begin{tabular}{c|c|c|c|c|cccc}
    \toprule
     % & $Prompt$ & $Enhance$ & $Img$ & $Ins$ & Daytime Foggy & Night Clear & Night Rainy & Dusk Rainy \\ 
     \multirow{2}{*}{} & \multirow{2}{*}{$Prompt$} & \multirow{2}{*}{$Enhance$} & \multirow{2}{*}{$Img$} & \multirow{2}{*}{$Ins$}  & Daytime & Night & Night & Dusk \\
     &  &  &  &  & Foggy & Clear & Rainy & Rainy \\
    \midrule 
1. & - & \checkmark & \checkmark & \checkmark & 37.8 & 38.3 & 17.1 & 32.2 \\
2. & \checkmark & -  & \checkmark & \checkmark & 38.5 & 38.7 & 16.9 & 32.8 \\
3. & \checkmark & \checkmark  & - & \checkmark & 39.2 & 39.8 & 18.5 & 33.1 \\
4. & \checkmark & \checkmark  & \checkmark & - & 38.9 & 39.5 & 18.3 & 32.8 \\
5. & \checkmark &  \checkmark  & - & - & 35.1 & 36.2 & 16.7 & 30.5 \\
6. & - &  \checkmark  & \checkmark & - & 35.6 & 36.1 & 16.3 & 29.9 \\
7. & - &  \checkmark  & - & \checkmark & 35.9 & 36.5 & 16.8 & 30.5 \\
8. & \checkmark &  -  & - & \checkmark & 32.9 & 34.8 & 14.2 & 27.5 \\
9. & \checkmark &  -  & \checkmark & - & 32.8 & 34.3 & 14.1 & 27.2 \\
10. & \checkmark &  -  & - & - & 32.1 & 34.0 & 13.5 & 26.5 \\
11. & \checkmark & \checkmark  & \checkmark & \checkmark & \textbf{40.0} & \textbf{41.5} & \textbf{19.8} & \textbf{34.5} \\
    \bottomrule
  \end{tabular}
  }
  \caption{Ablation study of internal modules. $Prompt$ denotes the proposed domain adaptation prompt, $Enhance$ represents the unified representation enhancement, $Img$ indicates the relative domain distance strategy, and $Ins$ signifies the positive-negative separation strategy.}
  % \vspace{-2mm}
  \label{tab:Ablation study of internal modules}
\end{table}

\begin{table}
  \centering
  \belowrulesep=0pt
  \aboverulesep=0pt
  \setlength{\heavyrulewidth}{1.5pt}
  \resizebox{8.3cm}{!}{
  \begin{tabular}{c|c|c|cccc}
    \toprule
    Size & $\mathcal{E}_{\mu}$ & $\mathcal{E}_{\sigma}$ & Daytime Foggy & Night Clear & Night Rainy & Dusk Rainy\\ 
    \midrule
     1 $\times$ 1 & - & \checkmark & 36.0 & 36.7 & 16.5 & 30.9 \\
     1 $\times$ 1 & \checkmark & \checkmark & 36.7 & 37.7 & 17.1 & 31.5 \\
     M $\times$ N & - & \checkmark & \underline{38.0} & \underline{38.5} & \underline{18.1} & \underline{33.3} \\
     M $\times$ N & \checkmark & \checkmark & \textbf{40.0} & \textbf{41.5} & \textbf{19.8} & \textbf{34.5} \\
    \bottomrule
  \end{tabular}
  }
  \caption{Ablation study of enhancements. To fit the feature size of $\mathcal{I}$, 
The sizes of {M, N} are set to {7, 7} under diverse weather conditions.}
  \label{tab:Ablation study of enhancements}
\end{table}

\noindent \textbf{Choice of Enhancements.}
One of the key factors of our method is the alteration of feature style through enhancement, enabling the acquisition of pseudo target domain features.
In Tab.~\ref{tab:Ablation study of enhancements}, we analyze various enhancement selections.
Compared to previous method\cite{fahes2023poda} that create pseudo target domain features at image level (Size 1$\times$1), our region-level design achieves superior results, with average 3.2\% mAP improvement.
Furthermore, compared with only applying $\mathcal{E}_{\sigma}$ \cite{vidit2023clip} to features, our approach achieves mAP improvements of 0.7\% and 2.1\% on 1 $\times$ 1 and M $\times$ N settings, respectively.

\begin{table}
  \centering
  \setlength{\belowcaptionskip}{-0.1cm}
  \belowrulesep=0pt
  \aboverulesep=0pt
  \setlength{\heavyrulewidth}{1.5pt}
  \resizebox{8.5cm}{!}{
  \begin{tabular}{l|c|cc}
    % \toprule
    % \multicolumn{3}{c}{Method} & \multicolumn{5}{c}{mAP} \\
    \toprule
     Method & Prior & Daytime Foggy & Night Rainy\\ 
    \midrule
    Gaussian & noise physics & 34.5 & 16.3 \\
    CLIP-GAP\cite{vidit2023clip} & static prompt & 36.9 & 18.7 \\
    PODA* \cite{fahes2023poda} & static prompt & \underline{39.2} & 19.0 \\
    DAI-Net*\cite{du2024boosting} & darkness physics & 36.7 & 18.9 \\
    PDD\cite{li2024prompt} & static prompt & 39.1 & \underline{19.2} \\
    UPRE & unbias prompt &\textbf{40.0} & \textbf{19.8} \\
    \bottomrule
  \end{tabular}
  }
  \caption{Comparison of various prompt priors}
  \label{tab:Comparison of various prompt priors}
\end{table}
\noindent \textbf{Different Prompt Priors.}
To investigate the effectiveness of different prompt priors, we studied four prior methods in Tab.~\ref{tab:Comparison of various prompt priors}, including noise physics, static prompt, dark physics, and our proposed unbiased prompt.
We use Gaussian noise as the noise physics with our framework. For the static prompt-driven results, we report the performance of CLIP-GAP \cite{vidit2023clip} , PODA*\cite{fahes2023poda} and PDD \cite{li2024prompt}.
Our approach achieves improvement of 4.5\% mAP  over noise physics methods and 1.2\% over static prompt methods.
We also report the results of the darkness physics prior method DAI-Net \cite{du2024boosting}.
DAI-Net*, a day-to-night adaptation method based on dark physics prompts, demonstrates excellent performance for Night Clear conditions but performs poorly in other scenarios.
In comparison to DAI-Net, our method achieves an average 2.1\% mAP gain, demonstrating that our method is applicable to various scenarios.

\begin{table}
  \centering
  \belowrulesep=0pt
  \aboverulesep=0pt
  \setlength{\belowcaptionskip}{-0.3cm}
  \setlength{\heavyrulewidth}{1.5pt}
  \resizebox{8.5cm}{!}{
  \begin{tabular}{c|c|cccc}
    % \toprule
    % \multicolumn{3}{c}{Method} & \multicolumn{5}{c}{mAP} \\
    \toprule
    Iterative train & Run Steps & Daytime Foggy & Night Clear & Night Rainy & Dusk Rainy\\ 
    \midrule
    - & - & \textbf{40.0} & \textbf{41.5} & \textbf{19.8} & \textbf{34.5} \\
    \checkmark & 100 & \underline{39.5} & \underline{38.8} & \underline{18.6} & \underline{33.9} \\
    \checkmark & 500 & 37.1 & 36.9 & 17.5 & 32.5 \\
    \checkmark & 1000 & 36.6 & 36.3 & 16.9 & 31.4 \\
    \bottomrule
  \end{tabular}
  }
  \caption{Training schedule of prompt and enhancement}
  \label{tab:Training schedule of prompt and enhancement}
\end{table}

\noindent \textbf{Training Schedule of Prompt and Enhancement.}
We investigate the impact of different training strategies, as illustrated in Tab.~\ref{tab:Training schedule of prompt and enhancement}. 
In the unified prompt and representation enhancement training stage, jointly training prompt representations and enhancement features demonstrates the best performance. 
As the interval step increases, a noticeable decline in model performance is observed.
The unified training of prompt and enhancement representations provides positive interaction.
Freezing one of these variables disrupts the unified nature of the training process, resulting in a suboptimal approach that is insufficient for mitigating either detection bias or domain bias.

\noindent \textbf{Influence of Instance-level Enhancement.}
Tab.~\ref{tab:Influence study of the PNS} highlights the effectiveness of our PNS strategy in improving performance.  
By separating positive and negative proposals and computing foreground ($\mathcal{L}_c$) and background ($\mathcal{L}_{\textsmaller{bg}}$) losses independently, our method achieves a substantial mAP gain, e.g., +1.6\% on Night Clear.  
Notably, $\mathcal{L}_{\textsmaller{bg}}$ performs competitively, reducing mAP by average 1.0\% compared to the vanilla cross-entropy loss ($\mathcal{L}_{ce}$).  
Moreover, $\mathcal{L}_c$ alone achieves comparable performance to the combined loss ($\mathcal{L}_{\textsmaller{bg}} + \mathcal{L}_c$), underscoring the importance of effectively modeling background context.

\begin{table}
  \centering
  \setlength{\abovecaptionskip}{0.2cm}
  \setlength{\belowcaptionskip}{0.0cm}
  \setlength{\heavyrulewidth}{1.5pt}
  \belowrulesep=0pt
  \aboverulesep=0pt
  \resizebox{8.5cm}{!}{
  \begin{tabular}{c|cccc}
    % \toprule
    % \multicolumn{3}{c}{Method} & \multicolumn{5}{c}{mAP} \\
    \toprule
    Methods & Daytime Foggy & Night Clear & Night Rainy & Dusk Rainy\\
    \midrule
    $\mathcal{L}_{ce}$ & 37.0 & 37.3 & 16.5 & 30.8 \\
    $\mathcal{L}_{ \textsmaller{{bg}}} $ only & 36.6 & 35.7 & 15.5 & 29.6 \\
    $\mathcal{L}_c$ only & \underline{37.6} & \underline{37.8} & \underline{17.1} & \underline{31.4} \\
    $\mathcal{L}_{ \textsmaller{{bg}}} $ + $\mathcal{L}_c$ & \textbf{38.7} & \textbf{38.9} & \textbf{17.8} & \textbf{32.8} \\
    \bottomrule
  \end{tabular}
  }
  \caption{Influence study of the PNS. $\mathcal{L}_{ce}$ is cross-entropy loss.}
  % \vspace{-2mm}
  \label{tab:Influence study of the PNS}
\end{table}

\begin{table}
  \centering
  \setlength{\abovecaptionskip}{0.1cm}
  \setlength{\belowcaptionskip}{-0.3cm}
  \setlength{\heavyrulewidth}{1.5pt}
  \belowrulesep=0pt
  \aboverulesep=0pt
  \resizebox{8.3cm}{!}{
  \begin{tabular}{l|c|c|c|cccc}
    \toprule 
    \multirow{2}{*}{Method} & Prompt & Category & Label of  & Daytime & Night & Night & Dusk \\
    & Design & Space & Negatives & Foggy & Clear & Rainy & Rainy \\
    \midrule
    DetPro & Negative & $\mathcal{C} \cup \textsmaller{\mathcal{C}_{bg}}$ & 0 or 1 & 38.1 &\underline{38.4} & \underline{19.0} & 32.8 \\
    DetPro & Shared & $\mathcal{C}$ & $\frac{1}{|\mathcal{C}|}$ & \underline{38.5} &{37.9} & {18.8} & \underline{33.3} \\
    \midrule
    PNS (Ours) & Negative & $\mathcal{C} \cup \textsmaller{\mathcal{C}_{bg}}$ & $\frac{1}{|\mathcal{C} \cup \textsmaller{\mathcal{C}_{bg}}|}$ & \textbf{40.0} & \textbf{41.5} & \textbf{19.8} & \textbf{34.5} \\
    \bottomrule
  \end{tabular}}
  \caption{DetPro vs. PNS: comparative analysis and ablation study.}
  \label{table2}
\end{table}

\begin{table*}
  \centering
  \setlength{\abovecaptionskip}{0.2cm}
  \setlength{\belowcaptionskip}{-0.1cm}
  \setlength{\heavyrulewidth}{1.2pt}
  \belowrulesep=0pt
  \aboverulesep=0pt
  \resizebox{16.3cm}{!}{
  \begin{tabular}{l|c|c|c|c|c|cccc}
    \toprule 
    \multirow{2}{*}{Method} & \multirow{2}{*}{Category} & Inference  & Computation & Total & Daytime & Daytime & Night & Night & Dusk \\
    &  & Time & Cost & Parameters & Clear & Foggy & Clear & Rainy & Rainy \\
    \midrule
    Faster-RCNN & \multirow{4}{*}{Traditional} & 67ms & 183G & 52.7M & 48.1 & 32.0 & 34.4 & 12.4 & 26.0 \\
    OA-DG &  & 69ms & 183G &52.7M & \underline{55.8} & 38.3 & 38.0 & 16.8 & \underline{33.9} \\
    DAI-Net* &  & 124ms & 297G & 78.1M & 54.4 & 36.7 &\underline{41.0} & 18.9 & 33.0 \\
    UFR &  & - & - & - &  \textbf{58.6} & \underline{39.6} &{40.8} & \underline{19.2} & 33.2 \\
    \midrule
    % S-DGOD & CVPR'22 & - & \underline{56.1} & 33.5 & 36.6 & 16.6 & 28.2 \\
    CLIP-GAP & \multirow{3}{*}{VLM-based} & 98ms & 526G &131.4M & 51.3 & 38.5 & 36.9 & 18.7 & 32.3 \\
    PODA* &  & 72ms & 185G & 129.3M & 51.8 & 39.2 & 38.7 & 19.0 & 33.4 \\
    PDD &  & 101ms & 531G & 134.3M & 53.6 & 39.1 & 38.5 & \underline{19.2} & 33.7 \\
    \midrule \midrule
    UPRE(Ours) & VLM-based  & 111ms & 528G & 129.8M & 53.9 & \textbf{40.0} & \textbf{41.5} & \textbf{19.8} & \textbf{34.5} \\
    \bottomrule
  \end{tabular}}
  \caption{The comparison of efficiency and effectiveness. All test settings are same. CLIP-GAP, PDD and UPRE are all based on the Detectron2 framework, using CLIP's ResNet101 backbone. UFR's - denotes code and model remain unavailable. PODA only use visual encoder during inference.}
  \label{better domain}
\end{table*}

\begin{figure*}
  \centering
  \setlength{\abovecaptionskip}{0.2cm}
  \setlength{\belowcaptionskip}{-0.2cm}
  \setlength{\heavyrulewidth}{1.5pt}
   \includegraphics[width=0.98\linewidth]{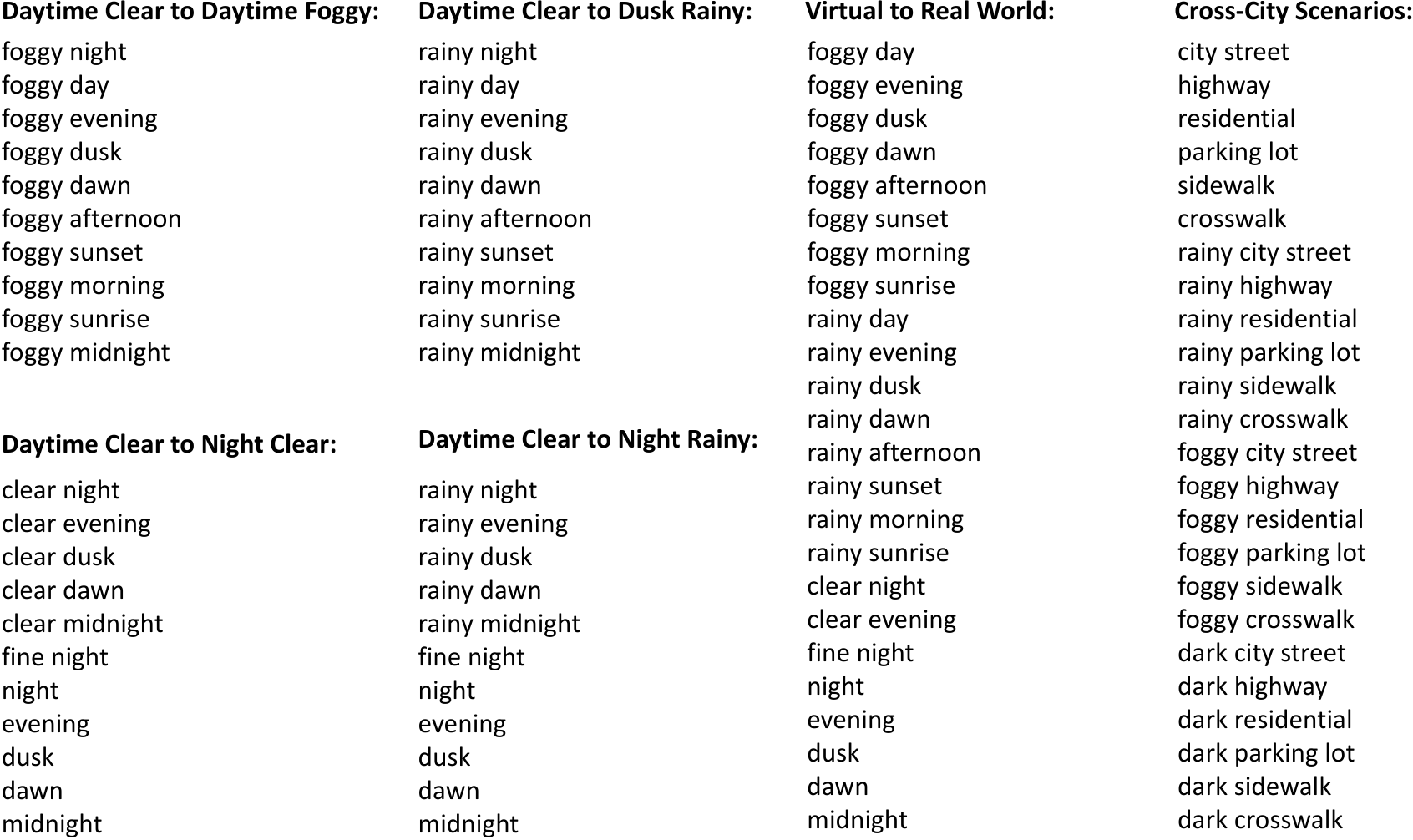}
   \caption{The domain prompt templates used for zero-shot domain adaptation object detection.
   }
   \label{prompt}
\end{figure*}

\begin{figure*}[t]
  \centering
  \setlength{\abovecaptionskip}{0.2cm}
  \setlength{\belowcaptionskip}{-0.2cm}
   \includegraphics[width=0.95\linewidth]{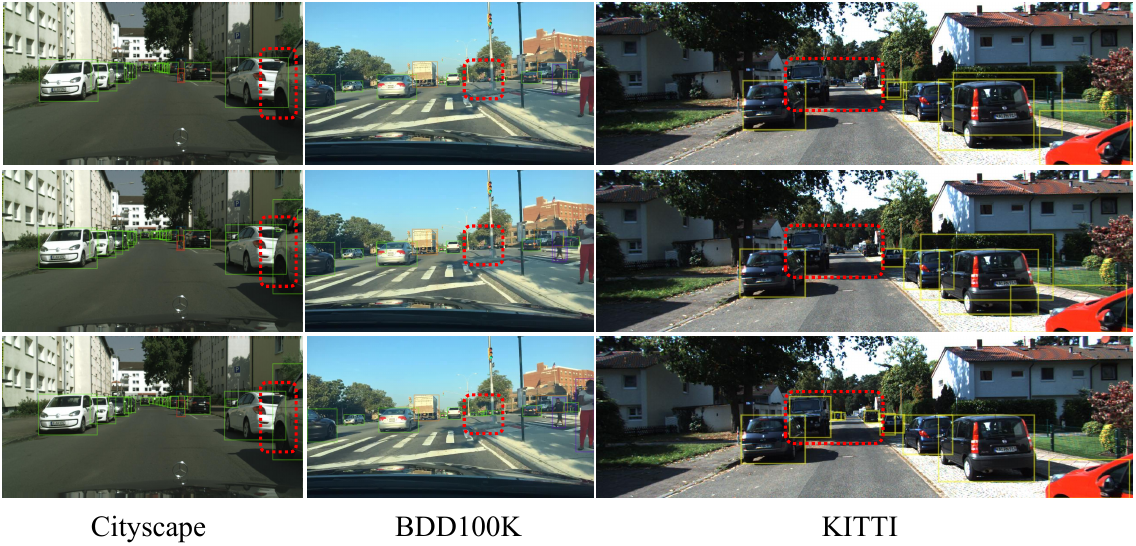}

   \caption{Visualization results under cross-city scenarios. The top to bottom rows show results from CLIP-GAP~\cite{vidit2023clip}, OA-DG~\cite{lee2024object}, and UPRE, respectively. 
}
   \label{cross-city}
\end{figure*}

\begin{figure*}
  \centering
  \setlength{\abovecaptionskip}{0.2cm}
  \setlength{\belowcaptionskip}{-0.2cm}
   \includegraphics[width=0.98\linewidth]{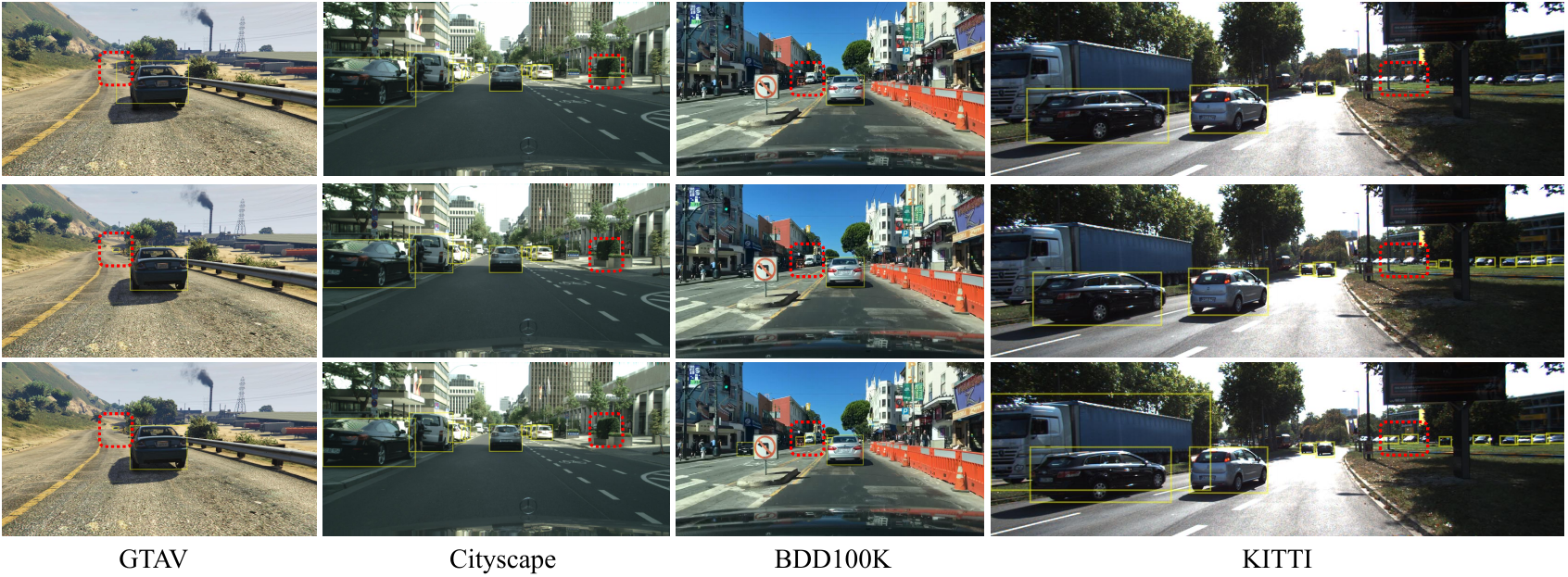}
   \caption{Visualization results of virtual-to-real world transitions. The top to bottom rows show results from CLIP-GAP~\cite{vidit2023clip}, OA-DG~\cite{lee2024object}, and UPRE, respectively. 
   }
   \label{virtual}
\end{figure*}

\noindent \textbf{Comparison with DetPro.}
In Table~\ref{table2}, we present ablation studies that highlight differences between PNS and DetPro.
As shown in line 1, DetPro struggles with training negative prompt, because it only learns a prompt embedding that draws all negative proposals close with a simple label of 0 or 1 (Detpro's Eq.(8)).
Therefore, as depicted in line 2, DetPro opts to train the shared prompt by forcing negative proposals to be equally unlike any \textbf{foreground classes} in category space $\mathcal{C}$ (Detpro's Eq.(5)).
However, selecting from foreground classes with a probability of $\frac{1}{|\mathcal{C}|}$ overlooks the utilization of the background category.

In contrast to DetPro, PNS trains negative prompt in category space $\mathcal{C} \cup \textsmaller{\mathcal{C}_{bg}}$, using $\frac{1}{|\mathcal{C} \cup \textsmaller{\mathcal{C}_{bg}}|}$ for labeling negative proposals in Eq.(12).
This approach allows negative proposals to be equally unlike any \textbf{classes} in Eq.(11), which aids the negative prompt in learning diverse context, as negative proposals variably encompass either pure backgrounds or parts of objects.
Contrasted with DetPro's shared and negative prompts, our method excels in performance across all target domains, achieving notable improvement by 3.6\% and 3.1\% in Night Clear, respectively.
DetPro insufficiently explores negative prompt, while PNS effectively trains the negative prompt.
PNS is not a direct application of DetPro; it innovatively explores proposal separation in training negative prompt.
Moreover, proposal separation is a commonly used trick in detection tasks.

\noindent \textbf{Efficiency analysis.}
Applying advanced large models~\cite{xiong2025hsstarhierarchicalsamplingselftaught,radford2021learning,li2022grounded} to ZSDA is promising but inevitably faces higher computational and memory costs.
As illustrated in Table~\ref{better domain}, UPRE outshines VLM-based methods, notably PDD, with 4.5M fewer parameters and a reduction of 3 GFLOPs in computational cost. Although UPRE is 10ms slower than PDD, this latency is acceptable given its superior performance.
To ensure efficiency, we only leverage the MDP module  during inference, which uses less than 0.1M parameters.

\noindent \textbf{Evidence of improvement attributed to domain adaptation.}
To assess whether improvements stem from better domain adaptation~\cite{he2025generalized,ji2025diffusion} or just a stronger detector, we compare different methods on the source domain to directly reflect detector performance. As shown in the Daytime clear column of Table \ref{better domain}, UPRE performs moderately but shows significant improvement in target domains.

\section{Prompt Templates \label{Prompt Engineering}}
\quad For a fair comparison, we adopt the same prompts used in CLIP~\cite{radford2021learning}.
Following previous works~\cite{vidit2023clip,li2024prompt}, we extend the category names and domain characteristics into sentences using multi-view prompts at both the image and instance levels.
Specifically, we use "A photo taken in a $[domain]$." as the image-level prompt input for the text encoder. 
Then, we define "A $[domain]$ photo of a $[class]$." as the instance-level positive prompt and "A $[domain]$ photo of an $[unknown \;\, class]$." as the instance-level negative prompt.
For the three cross-domain scenarios, we employ 90 $[domain]$-specific prompt templates, as illustrated in Figure~1. 
We apply L2 normalization to obtain the final multi-view prompt representations.

\section{More Implementation Details \label{More Training Details}} 
\quad In CLIP, the input is a $224 \times 224$ image, and the final Attention Pool processes a $7 \times 7$ feature map.
Current approaches~\cite{vidit2023clip} primarily rely on image cropping for data augmentation, resizing images to $224 \times 224$. However, this method conflicts with the nature of object detection, as images often contain multiple object instances. 
To ensure that training images retain more object instances, we use the original $1067 \times 600$ images as input. 
In the relative domain distance strategy, to align the feature size from the third layer of the CLIP image encoder with the input size of the CLIP Attention Pool, we first downsample the third-layer features to $21 \times 21$. Subsequently, a $3 \times 3$ average pooling layer is applied to produce a $7 \times 7$ feature map.
In the positive-negative separation strategy, the third-layer features from the CLIP image encoder serve as input to the RPN head. These features are then processed by ROI-Align to extract $14 \times 14$ region features. Next, the $14 \times 14$ region features are passed through the fourth layer of the image encoder, resulting in $7 \times 7$ detection features. Finally, these features are fed into the classification and bounding box regression heads to generate the detection results.

\section{Additional Visualizations \label{More Visualization}}
\quad In this section, we provide more visualization results
under cross-city scenarios and virtual-to-real world transitions.
As shown in Fig.~\ref{cross-city}, our method achieves the best performance, while other methods exhibit issues with both duplicate detections and miss detections, demonstrating the effectiveness of our approach.
Compared to cross-city scenarios, the virtual-to-real scenario is more challenging due to the significant domain gap between the synthetic GTAV game world and the real world.
Despite this challenge, our method achieves satisfactory detection accuracy compared to other approaches (see Fig.~\ref{virtual}).
This validates the theory that incorporating real-world weather styles into a clear virtual environment can effectively transform it into a realistic representation under various weather conditions.
\end{document}